\documentclass[11pt]{article}

\usepackage[final]{acl}

\usepackage{times}
\usepackage{latexsym}
\usepackage{booktabs}

\usepackage[T1]{fontenc}

\usepackage[utf8]{inputenc}

\usepackage{microtype}
\usepackage{hyperref}
\usepackage{stfloats}   

\usepackage{inconsolata}

\usepackage{graphicx}
\usepackage{float}
\usepackage{needspace}
\usepackage[most]{tcolorbox}
\usepackage{amsmath}
\usepackage{amssymb}
\usepackage{tikz}
\usepackage{xcolor}
\usepackage{colortbl}
\usepackage{xspace}

\newcommand{\AS}{{\sc Hero's Journey}\xspace}
\DeclareRobustCommand{\cnum}[1]{%
  \tikz[baseline=(c.base)]{%
    \node[circle,fill=black,inner sep=1pt,
          font=\scriptsize\bfseries\color{white}] (c) {#1};%
  }%
}

\tcbset{
  base/.style={
    arc=0mm,
    bottomtitle=0.5mm,
    boxrule=0mm,
    colbacktitle=black!10!white,
    colframe=black!30!white,
    coltitle=black,
    fonttitle=\bfseries\fontsize{10}{12}\selectfont,
    left=2.5mm,
    leftrule=1mm,
    right=3.5mm,
    title={#1},
    toptitle=0.75mm,
    breakable,
    listing options={style=tcblatex, breakindent=0pt, breaklines=true,
                     basicstyle=\ttfamily\fontsize{8}{9}\selectfont}
  }
}

\newtcblisting[use counter=codeboxinput]{codebox}[2][]{
  base={Listing \thetcbcounter: #2},
  listing only,
  #1
}
\newtcbinputlisting[use counter=codeboxinput]{\codeboxinput}[3][]{%
  listing file={#3},
  base={Listing \thetcbcounter: #2},
  listing only,
  #1
}
\newtcolorbox{subbox}[2][]{
  colframe=black!30!white,
  base={#2},
  #1
}

\usepackage[textsize=scriptsize]{todonotes}

\title{\raisebox{-4pt}{\includegraphics[height=25pt]{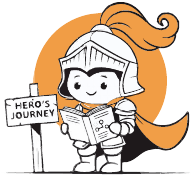}}~\AS: Testing Complex Rule Induction with Text Games}

\author{Anshun Asher Zheng, Kanishka Misra, David I. Beaver, Junyi Jessy Li\\ Department of Linguistics\\
        The University of Texas at Austin \\ \texttt{\{asher.zheng, kmisra, dib, jessy\}@utexas.edu}}

\begin{document}
\maketitle
\begin{abstract}
We introduce \textbf{\AS}, a benchmark for rule induction in goal-directed episodic tasks, where agents must infer hidden rules from demonstrations and act on them through multi-step execution.
\AS\footnote{We release the \href{https://github.com/asherz720/HerosJourney}{code} and the \href{https://pypi.org/project/herosjourney/0.1.0/}{package} under MIT license.} covers eight tasks across attribute and procedural induction families, each with four structural rule forms, controllable lexical grounding, and identifiability conditions.
Evaluating state-of-the-art LLMs, we find that models show evidence of rule induction, but the ability is limited and uneven across tasks. Meanwhile, process execution adds an execution bottleneck for models, whereas surface semantics has minimal effect. Induction-specific steering methods improve performance on attribute tasks but show no reliable gains on procedural tasks, suggesting the gap in procedural induction remains an open challenge.
\end{abstract}

\section{Introduction}

\begin{figure*}[t]
  \centering
  \includegraphics[width=\textwidth]{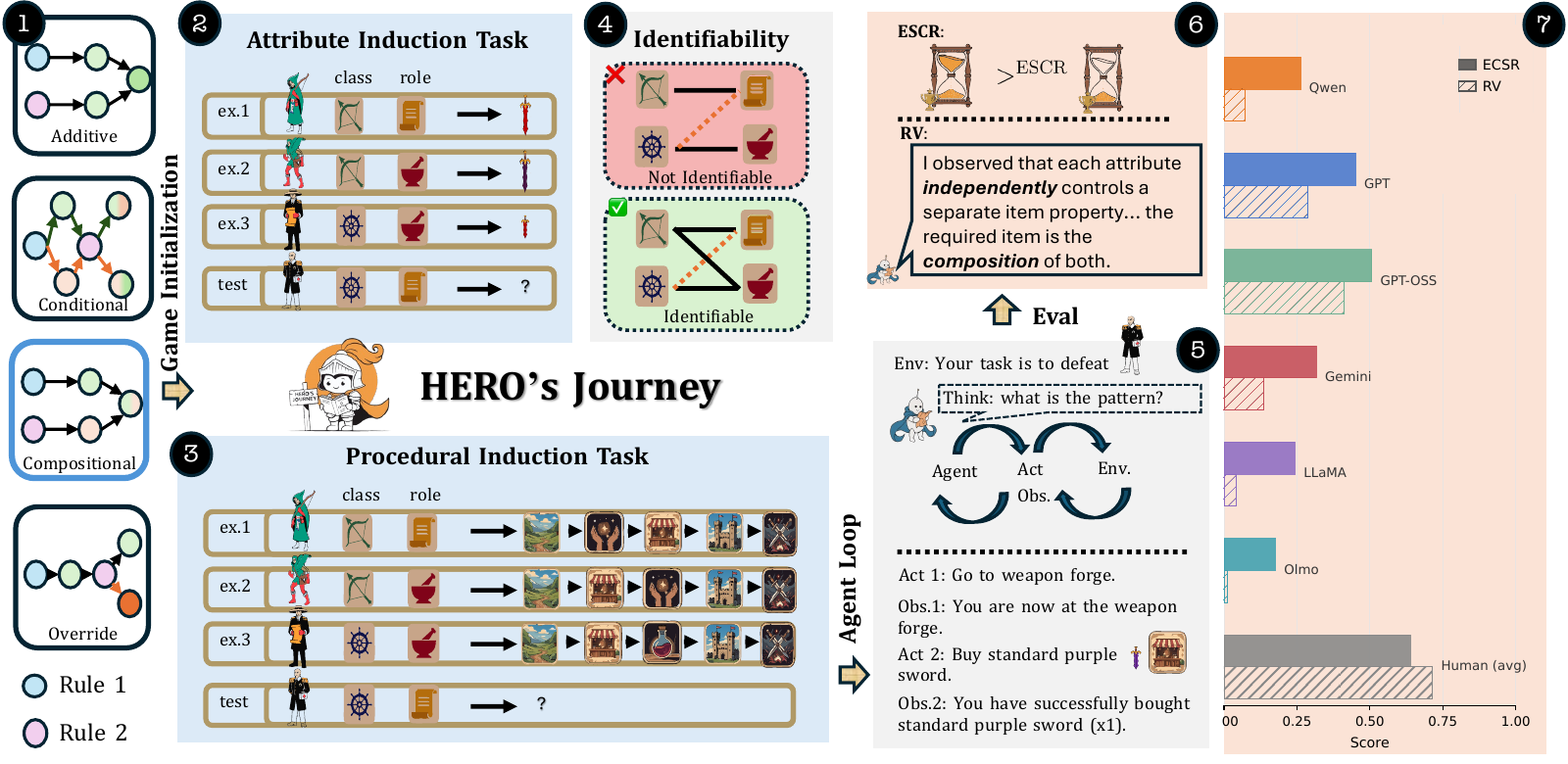}
  \caption{%
  \textbf{Overview of \AS.}
  \cnum{1}~\textbf{Rule interactions:} four structural forms varying how entity attributes jointly determine the required item or process.
  \cnum{2}~\textbf{Attribute induction tasks:} illustrated with A-Comp tasks (§\ref{attribute induction}; each attribute independently governs a separate output dimension); entity class (\raisebox{-3pt}{\includegraphics[height=3ex]{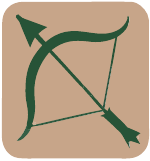}}~ranger,
  \raisebox{-3pt}{\includegraphics[height=3ex]{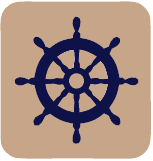}}~captain)
  and role (\raisebox{-3pt}{\includegraphics[height=3ex]{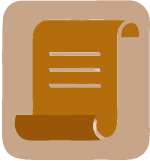}}~prophet,
  \raisebox{-3pt}{\includegraphics[height=3ex]{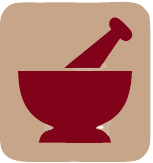}}~chirurgeon).
  \cnum{3}~\textbf{Procedural induction tasks:} illustrated with P-Comp tasks (§\ref{procedural induction}; same structure):
  \raisebox{-3pt}{\includegraphics[height=3ex]{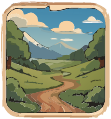}}~start,
  \raisebox{-3pt}{\includegraphics[height=3ex]{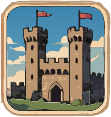}}~go\_to\_entity\_location,
  \raisebox{-3pt}{\includegraphics[height=3ex]{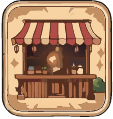}}~buy\_objects,
  \raisebox{-3pt}{\includegraphics[height=3ex]{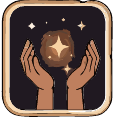}}~perform\_ritual,
  \raisebox{-3pt}{\includegraphics[height=3ex]{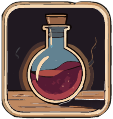}}~drink\_potion,
  \raisebox{-3pt}{\includegraphics[height=3ex]{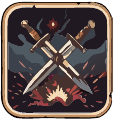}}~defeat\_entity.
  \cnum{4}~\textbf{Identifiability:} source demonstrations cover all combinations needed to recover the rule.
  \cnum{5}~\textbf{Agent loop:} the agent reads demonstrations and acts in the environment.
  \cnum{6}~\textbf{Evaluations:} ECSR measures task execution efficiency; RV measures whether the agent can verbalize the rule.
  \cnum{7}~\textbf{Results:} ECSR and RV across all models; all underperform the human baseline.
  }
  \label{fig:overview}
\end{figure*}

Somewhere beyond the dark forest, a captive waits to be freed.
A warrior sets out, brave, but the only knowledge they carry is what others have left behind: scattered accounts of warriors who walked similar paths before, each facing a different guardian, each prevailing with a different weapon.
The castle gates will not open until the guardian falls, and the guardian will not fall unless the warrior arrives with the right weapon.
There is no oracle to consult; the armory is a day's ride away and a wrong choice means starting over.
The warrior must read the old accounts, find the pattern hiding in them, and commit. This simple scenario captures two challenges that, together, define the problem we study.
The first is \textbf{inductive}: the rule connecting a guardian's traits to the weapon that defeats it is never stated and must be recovered from past episodes.
The second is \textbf{procedural}: acting on the rule requires executing a sequence of steps in the right order, where an incorrect inference forces the warrior to start again.

From research agents that autonomously design and execute experiments \citep{lu2024aiscientist}, to coding agents that resolve real-world software issues \citep{jimenez2024swebench}, to web agents navigating complex multi-step tasks \citep{zhou2024webarena},
LLM agents must infer patterns from past examples and act on them. 
 Despite rapid capability gains, LLMs continue to struggle with inductive reasoning \citep{li2025mirage}. For example, models in rule-learning settings frequently default to surface retrieval rather than structural induction \citep{he2024idea}. Existing benchmarks leave three aspects underexplored: (i) induction is not always situated in a multi-step execution setting where the inferred rule must drive a sequence of dependent actions (cf. \citealt{chollet2019arc}); (ii) evaluations typically test atomic mappings in isolation rather than more complex interaction between rules (cf. \citealt{rule2020child}); and (iii) demonstrations may leave multiple rules consistent with the evidence, so it is unclear whether evaluation targets the specific rule under
investigation (cf. \citealt{qiu-etal-2024-phenomenal, he2024idea}). We refer interested readers to \citet{chen2025survey} for a systematic review.

To address these gaps, we introduce \textbf{\AS}, an evaluation suite for rule induction in goal-directed episodic tasks.
In each episode, the agent must infer the rules from demonstration episodes and apply the induced rule to novel test entities by executing a sequence of actions in the right order (Fig. \ref{fig:overview} \cnum{5}). We instantiate \AS with two families of tasks.
\textit{Attribute induction} tasks (Fig. \ref{fig:overview} \cnum{2}; \S\ref{attribute induction}) require inferring a hidden mapping from entity attributes to required items, with the mapping varying across four structural forms: additive, compositional, conditional, and override, each representing a different interaction between rules (Fig. \ref{fig:overview} \cnum{1}).
\textit{Procedural induction} tasks (Fig. \ref{fig:overview} \cnum{3}; \S\ref{procedural induction}) require inferring hidden steps or ordering constraints within the goal procedure, with analogous structural variation. Each task comes with an identifiability condition (Fig. \ref{fig:overview} \cnum{4}) that ensures the demonstration episodes are collectively sufficient to recover the target rule.

Performance on \AS is evaluated with two metrics for rule induction: ECSR (efficiency-calibrated success rate) captures whether models succeed efficiently (i.e., not via brute-force enumeration), and RV (rule verbalization) captures whether they can explicitly articulate the test rules. (Fig. \ref{fig:overview} \cnum{6}). 
Across frontier proprietary and open models, humans outperform the strongest model on average by 13\% on ECSR and 30\% on RV, though the strongest models approach human ECSR on select attribute tasks (Fig. \ref{fig:overview} \cnum{7}).
While models are showing some reliance on rule induction to achieve tasks, the ability is still very limited and models may instead rely on contextual bias to succeed (\S\ref{sec:rule_induction}).

Notably, by embedding the tasks in a multi-step execution setting, surface semantics alone would not be effective. This extra complexity reveals further gaps in existing methodologies: steering methods yield only limited improvement and remain far below human-level performance   
(\S\ref{sec:other_factors}).



\section{Background and Design Principles}
\label{background}

\paragraph{Inductive reasoning and its evaluation in LLMs.}
Inductive reasoning, deriving general rules from specific observations and applying them to novel instances, is a core capacity of intelligent agents \citep{tenenbaum2011grow, lake2015human}. A related tradition in cognitive science has studied \textit{property induction}: how agents generalize properties across entities sharing common category membership \citep{Osherson1990CategoryBasedI, Sloman1993FeatureBasedI, Kemp2013ATO, Hayes2018InductiveR2}; language models have been also evaluated on tasks of this kind \citep{ misra2022property, bhatia2024inductive, han2024inductive, padmanabhan2025language}. More recently, it is found that LLMs in some settings perform explicit rule induction \citep{he2024idea, lee-etal-2025-moc}, though \citet{li2025mirage} reveal a systematic dissociation between the inductive and deductive stages: models often arrive at correct outputs without having abstracted the underlying rule.

Existing benchmarks have probed abstract pattern induction across sequences, lists, and visual grids \citep{chollet2019arc, rule2020child, zhang2021acre}, and compositional generalization benchmarks have revealed systematic failures when applying learned rules to novel combinations \citep{lake2018generalization, keysers2020measuring, kim-linzen-2020-cogs, ruis2020benchmark}. 

However, existing evaluation settings leave important dimensions of rule induction underexplored. First, neither symbolic nor compositional benchmarks embed inductive reasoning in a \textit{procedural task-execution setting}: evidence comes from static input-output pairs rather than action trajectories, and the induced rule need not drive planning within a multi-step procedure \citep[cf.][]{lake2018generalization, chollet2019arc}; an exception is \citet{he2024idea}, which uses a text-based game environment, though it focuses on self-directed exploration of the rule rather than induction from fixed demonstrations. Existing benchmarks typically test atomic mappings (i.e., single input-output rules) in isolation, leaving open how models handle more complex rule interactions \citep[cf.][]{rule2020child}. Importantly, evaluation settings that allow models to freely explore and accumulate evidence do not enforce a principled identifiability condition: multiple rules may remain consistent with the observed evidence, leaving it unclear whether the rule under test was induced \citep[cf.][]{qiu-etal-2024-phenomenal, he2024idea}.

\paragraph{\AS Desiderata.}
The above gaps matter because they leave open the very questions we care about: whether a model that identifies a rule can actually act on it, whether observed successes reflect the target rule or merely a consistent alternative, and whether inductive capacity extends beyond the simplest atomic mappings. 
We therefore identify three desiderata in our design:

\textbf{(1)} \textit{Evaluation should be embedded in a procedural task-execution setting}, so that the agent must not only induce the rule but act on it through dependent steps, revealing whether induction translates to correct behavior, not just correct answers.

\textbf{(2)} \textit{Rule interaction should be configurable}: varying not just what the rule maps but how rules interact allows evaluation to go beyond the atomic mappings that current benchmarks test in isolation.

\textbf{(3)} \textit{Demonstrations should satisfy an identifiability condition}: source examples must sufficiently determine the target rule so that tasks are well-posed and task performance can be more reliably attributed to the induction of the target rule.

Text-based game environments \citep{cote2018textworld, chevalier-boisvert2019babyai,hausknecht2020interactive} provide a natural platform for all three: they support goal-directed multi-step action in a fully synthetic world where rule structure and source/gen splits can be independently configured; the ability to swap naturalistic names for surface-neutral alternatives further allows pretraining associations to be disentangled from structural induction \citep{min-etal-2022-rethinking}. 


\section{\AS}
\label{AS}

\textbf{\AS} is a controlled evaluation suite for rule induction in goal-directed episodic tasks, built on a deterministic text-based environment where an agent pursues goals via natural-language actions and certain goal requirements are omitted and must be inferred from demonstrations.\footnote{We instantiate \AS as an RPG-style adventure domain, but the framework transfers to other domains by defining a new action set and lexicon, leaving rule structure and evaluation logic unchanged.}

\subsection{Task Curation Process}

\begin{figure*}[t]
  \centering
  \includegraphics[width=\textwidth]{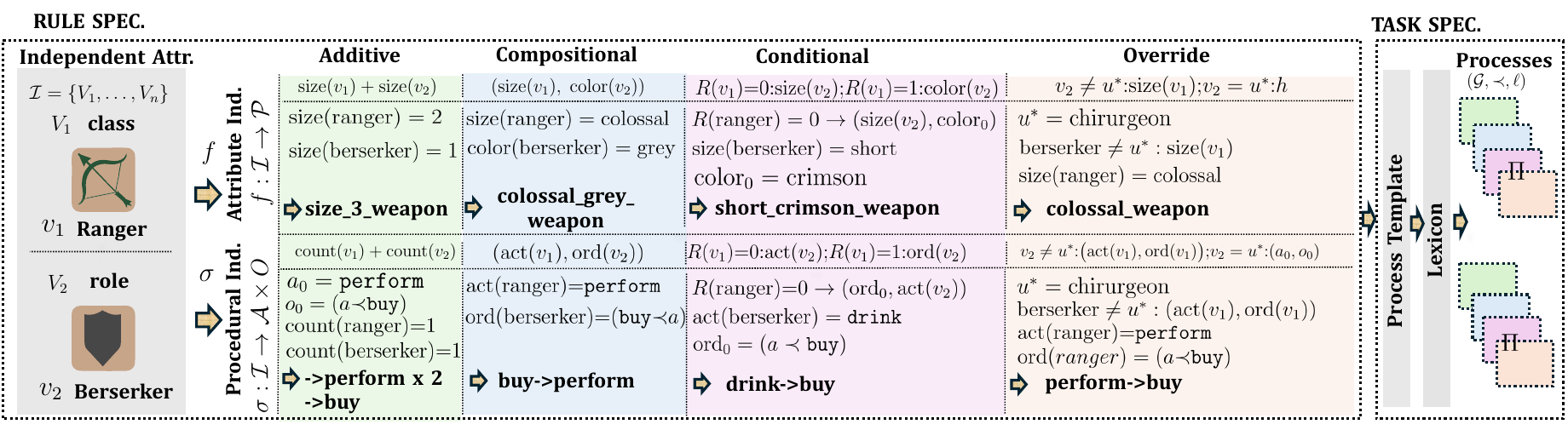}
  \caption{Task curation and the eight induction tasks.
  \textit{Left} (Rule Spec.): independent attributes $\mathcal{I}$ ($v_1$\,=\,ranger, $v_2$\,=\,berserker as a concrete example).
  \textit{Middle}: structural forms of $f{:}\mathcal{I}{\to}\mathcal{P}$ (top, Attr1--4) and $\sigma{:}\mathcal{I}{\to}\mathcal{A}{\times}O$ (bottom, Proc1--4).
  \textit{Right} (Task Spec.): rule function outputs are combined with the process template to populate each task type; see also Figure \ref{fig:task_curation} (Appx.~\ref{appendix:task_details}) for a concrete illustration.}
  \label{fig:task_overview}
\end{figure*}

Each task in \AS is a set of \textit{processes} curated from a \textit{rule specification}, and a \textit{task specification} (Figure~\ref{fig:task_overview}). To make the curation process concrete, we develop it through a running example. A wandering warrior must defeat the guardian of the Dragon's Keep. The keep's records list the guardian's class and role, but say nothing of what can defeat it
; the only clues are tales from adventurers who walked this path before.

\noindent\textbf{Process.}~Studying the accounts, the warrior notices a shared pattern: stop at an armory, acquire a weapon, travel to the keep, face the guardian. The sequence of actions an agent must perform to succeed is what a process represents. Formally, a process $\Pi$ is a goal network $(\mathcal{G}, \prec, \ell)$ where $\mathcal{G}$ is a finite set of subgoals plus a terminal goal 
(each written $g_i \in \mathcal{G}$), $\prec$ is a strict partial order over $\mathcal{G}$ encoding prerequisite dependencies, and $\ell: \mathcal{G} \to \mathcal{A} \times \mathcal{X}$ assigns each node an action $a \in \mathcal{A}$ and an argument $x \in \mathcal{X}$ (i.e., the specific object or location the action acts on; e.g., \texttt{go}(\textit{location}), \texttt{buy}(\textit{weapon})). An agent completes an episode by achieving all subgoals
in an order consistent with $\prec$ and reaching the terminal goal. The process defines which actions are required and in what order, and what \textit{type} of argument each accepts, but not the specific values those arguments take; keeping the two separate allows procedural and attribute-level generalization to be controlled and evaluated independently.

\noindent\textbf{Rule Specification.}~The weapon choice is not arbitrary: it depends on the guardian's class and role. More generally, what an entity requires is determined by its attributes via a hidden \emph{rule function}. We call class and role the \emph{independent} attributes ($\mathcal{I}$): they are observable and serve as input to the rule. The codomain differs by task family. For \textbf{attribute induction}, the rule $f$ maps attribute assignments to item properties (e.g., weapon size or color). For \textbf{procedural induction}, the rule $\sigma$ maps them to a process variant, specifying which action to insert and where relative to a reference node. The structural form of $f$ (resp.\ $\sigma$) is the primary design variable across the tasks in \S\ref{attribute induction}--\ref{procedural induction}.



\noindent\textbf{Task Specification.}~A task specification individuates each task: it declares which rule function applies (i.e., $f$, $\sigma$), where they are applied (i.e., the base process templates), and how the source/gen split is constructed to satisfy identifiability (Appx.~\ref{appx:identifiability}). A lexicon then instantiates all abstract attribute values and outcomes into surface names---\emph{semantic} (e.g., \emph{dragon-blooded}) or \emph{nonce} (e.g., \emph{krev\_A})---isolating structural induction from lexical world knowledge (see details in Appx.~\ref{appx:task_generation}).


\subsection{Inductive Reasoning Tasks with \AS}
We instantiate \AS with tasks targeting two types of generalization. \textbf{Attribute induction} tasks require the agent to infer a hidden mapping from entity attributes to required items and apply it to novel entities. \textbf{Procedural induction} tasks require the agent to infer hidden steps or ordering constraints needed to achieve a goal. We describe each below. Figure~\ref{fig:task_overview} (middle) illustrates a running example we use for each task; full instantiation details are in Appx.~\ref{appendix:task_details}.

\subsubsection{Attribute Induction Tasks (Attr1--4)}
\label{attribute induction}
\noindent\textbf{Formalization.}~In attribute induction tasks, the rule specification declares (1) independent attributes ($\mathcal{I}$) with value sets $V_1, \ldots, V_n$ and (2) dependent item properties $\mathcal{P}$ with value sets $W_1, \ldots, W_n$. The hidden mapping $f$, which the agent must induce, maps assignments over the entity attributes to item property. In our instantiation we use $n$ as 2, giving $f: V_1 \times V_2 \to \mathcal{P}$. The four tasks each instantiate a structurally distinct form of $f$, described below. To ground the tasks concretely, consider a ranger-class ($v_1$) guardian with a berserker role ($v_2$) in Dragon's Keep; the inferential challenge is which item to acquire to defeat the guardian.

\noindent\textbf{Attr1: Additive.}~\textit{A-Add} instantiates $f$ as $\text{size}(v_1)$ ${+} \text{size}(v_2)$, where $\text{size}{:}V_i{\to}\mathbb{Z}_{\geq 0}$ assigns a size integer to each attribute independently. In our instantiation, ranger contributes size~2 ($\text{size}(v_1){=}2$) and berserker contributes size~1 ($\text{size}(v_2){=}1$), so to defeat the guardian requires a size 3 weapon.

\noindent\textbf{Attr2: Compositional.}~\textit{A-Comp} instantiates $f$ as $(\text{size}(v_1),\,\text{color}(v_2))$, where $\text{size}{:}V_1{\to}W_\text{size}$ and $\text{color}{:}V_2{\to}W_\text{color}$ are independent. Here, a ranger class requires colossal size ($\text{size}(v_1){=}\text{colossal}$) and a berserker role requires grey color ($\text{color}(v_2){=}\text{grey}$), so to defeat the guardian requires a colossal-grey weapon.

\noindent\textbf{Attr3: Conditional.}~\textit{A-Cond} introduces a \emph{regime} function $R{:}V_1{\to}\{0,1\}$ that partitions $V_1$ into two groups, each determining which item property $V_2$ governs:
\[ f(v_1, v_2) = \begin{cases} \bigl(\text{size}(v_2),\;\text{color}_0\bigr) & \text{if } R(v_1)=0 \\ \bigl(\text{size}_0,\;\text{color}(v_2)\bigr) & \text{if } R(v_1)=1 \end{cases} \]
where $\text{size}_0$, $\text{color}_0$ are fixed per-regime constants. In our instantiation, ranger falls in regime~0 ($R(v_1){=}0$), so berserker governs size: $\text{size}(v_2)$ ${=}$ short with color fixed at crimson, yielding a short-crimson weapon. For a guardian of the captain class in regime~1 ($R(\text{captain}){=}1$), berserker instead governs color: $\text{color}(v_2){=}$crimson with size fixed at long, yielding a long-crimson weapon.


\noindent\textbf{Attr4: Override.}~\textit{A-Over} instantiates $f$ as a size mapping governed by $V_1$, unless $V_2$ matches a designated exception $u^* \in V_2$ that overrides all class-based structure:
\[ f(v_1, v_2) = \begin{cases} h & \text{if } v_2 = u^* \\ \text{size}(v_1) & \text{otherwise} \end{cases} \]
where $\text{size}{:}V_1{\to}W_\text{size}$ is the base mapping and $h$ is the fixed override item. In our instantiation, ranger paired with berserker (not $u^*$) follows the base rule: $\text{size}(v_1){=}$standard, yielding a standard weapon. But if the role is chirurgeon ($v_2{=}u^*$), the guardian always requires a great weapon ($h$) regardless of class.


\subsubsection{Procedural Induction Tasks (Proc1--4)}
  \label{procedural induction}
  \noindent\textbf{Formalization.}~In procedural induction tasks, the rule specification declares the same independent attributes $\mathcal{I} = V_1 \times V_2$ but the hidden rule $\sigma: \mathcal{I} \to \mathcal{A} \times O$ maps attribute assignments to process variants rather than item properties, where $\mathcal{A}$ is the action space and $O$ is the set of ordering positions relative to a reference node. Entities of different attributes require processes that vary in action type $a \in \mathcal{A}$, ordering $o \in O$, and (for the additive form) count $n$. The four structural forms of $\sigma$ mirror those of $f$. We use the same ranger-class ($v_1$) guardian with a berserker role ($v_2$) to ground each task concretely; the inferential challenge here is: \emph{which} action to take, \emph{how many} times, and \emph{where}.


\noindent\textbf{Proc1: Additive.}~\textit{P-Add} instantiates $\sigma$ as an additive count: the number of inserted nodes with fixed action $a_0\in\mathcal{A}$ at fixed ordering $o_0$ equals $\text{count}(v_1){+}\text{count}(v_2)$, where $\text{count}{:}V_i{\to}\mathbb{Z}_{\geq 0}$ assigns a node count to each attribute independently. In our instantiation, $a_0{=}$\texttt{perform} and $o_0{=}(a{\prec}\texttt{buy})$; ranger and berserker each contribute count~1, so the agent must execute \texttt{perform}${\to}$\texttt{perform}${\to}$\texttt{buy}.

\noindent\textbf{Proc2: Compositional.}~\textit{P-Comp} instantiates $\sigma$ as a product of two independent mappings: $\text{act}{:}V_1{\to}\mathcal{A}$ determines the action and $\text{ord}{:}V_2{\to}O$ determines the ordering. Ranger gives $\text{act}(v_1){=}$\texttt{perform} and berserker gives $\text{ord}(v_2){=}(\texttt{buy}{\prec}a)$, yielding \texttt{buy}${\to}$\texttt{perform}.

\noindent\textbf{Proc3: Conditional.}~\textit{P-Cond} instantiates $\sigma$ with the same regime structure as \textit{A-Cond}: $R{:}V_1{\to}\{0,1\}$ selects which knob $V_2$ governs: in regime~0, $V_2$ determines action (ordering fixed at $\text{ord}_0$); in regime~1, $V_2$ determines ordering (action fixed at $\text{act}_0{=}$\texttt{drink}). Ranger (regime~0) paired with berserker gives $\text{act}(v_2){=}$\texttt{drink} with ordering fixed at $(a{\prec}\texttt{buy})$, yielding \texttt{drink}${\to}$\texttt{buy}. For a guardian of the captain class (regime~1), berserker gives $\text{ord}(v_2){=}(\texttt{buy}{\prec}a)$ with action fixed at \texttt{drink}, yielding \texttt{buy}${\to}$\texttt{drink}.

\noindent\textbf{Proc4: Override.}~\textit{P-Over} instantiates $\sigma$ with the same exception structure as \textit{A-Over}: a designated $u^*{\in}V_2$ overrides both knobs to fixed $(a_0,o_0)$ regardless of $V_1$; otherwise $V_1$ governs via $\text{act}(v_1)$ and $\text{ord}(v_1)$. Ranger paired with berserker (not $u^*$) gives $\text{act}(v_1){=}$\texttt{perform} and $\text{ord}(v_1){=}(a{\prec}\texttt{buy})$, yielding \texttt{perform}${\to}$\texttt{buy}; chirurgeon ($v_2{=}u^*$) always overrides to $(a_0,o_0){=}($\texttt{drink}$,\texttt{buy}{\prec}a)$, yielding \texttt{buy}${\to}$\texttt{drink} regardless of class.

  \noindent\textbf{Process Template.}~To make both task families testable, we use a shared base episode template, where $f$ determines item selection and $\sigma$ determines process structure: go(\textit{item\_loc}) $\to$ buy(\textit{item}) $\to$ go(\textit{entity\_loc}) $\to$ defeat(\textit{entity}).\footnote{This particular sequence is the one used in our instantiation; there is no formal requirement to use it.}

\subsection{Identifiability Conditions}
\label{sec:identifiability}

We construct source demonstrations to satisfy an \emph{identifiability condition} for each task: the demonstrations contain enough signal to recover the target rule. This ensures our tasks are well-posed for studying rule induction and makes observed successes more interpretable (i.e., the target rule is at least a recoverable explanation). Take \textit{A-Comp} as a concrete illustration (see Figure~\ref{fig:overview} \cnum{4}): the rule $f(v_1,v_2)=\bigl(\text{size}(v_1),\,\text{color}(v_2)\bigr)$ maps each attribute to a different output dimension independently.
To recover the structure, the source split must provide: (1) pairs where one attribute varies while the other is held fixed to reveal independence (i.e., a bipartite-connected graph over $V_1\times V_2$); and (2) each attribute value appears at least once, so that every mapping entry is observable.
Without it, say the model only sees (\textit{ranger}, \textit{prophet})$\to$(\textit{large}, \textit{red}) and (\textit{captain}, \textit{chirurgeon})$\to$(\textit{short}, \textit{purple}), the setup would lead to a disconnected graph, leaving the contribution of each attribute 
ambiguous. For example, \textit{ranger} could map to \textit{large}, to \textit{red}, or to nothing at all if \textit{prophet} alone determines both dimensions.
We provide the full per-task conditions in Appx.~\ref{appx:identifiability}.
\begin{figure}[t]
  \centering
  \includegraphics[width=0.5\textwidth]{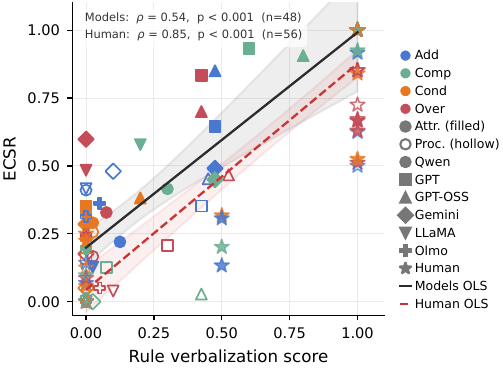}
  \caption{ECSR vs.\ RV across all model$\times$task conditions.
  Each point represents one model on one task.}
  \label{fig_ee_rv_combined}
\end{figure}
\section{Evaluation}
\label{sec:setup}

For each task in \S\ref{AS}, we generate 20 variants by randomly sampling surface names from the lexicon and varying the source-gen splits. In each episode the agent receives (see prompt in Appx.~\ref{appx:main-task-prompt}): (1) a \emph{world listing} of all entities, attributes, and locations; (2) source-split demonstrations from which the hidden rule can be inferred, interleaved with distractor trajectories $D$ that carry no underlying pattern; and (3) a gen-split goal for which the agent must execute the correct action sequence.  Each episode imposes an action budget of $n_\text{tries}$, the number of attempts a brute-force enumeration would require. A rule-inducing agent should solve the task well within this budget, while an exhaustive enumerator would just barely fit (see details in Appx.~\ref{appendix:eval_setup}).

\begin{figure*}[ht]
  \centering
  \includegraphics[width=\textwidth]{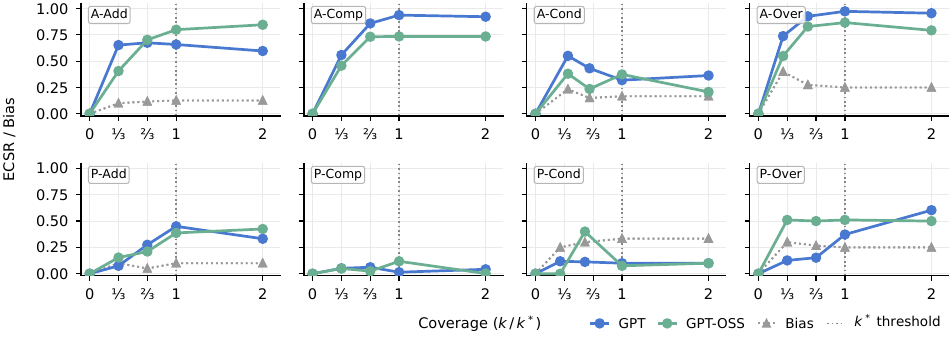}
  \caption{ECSR (solid colored lines) and
  $\textnormal{contextual\_bias}(k)$ (dotted gray triangles) across different
  coverage $k/k^*$, for all eight tasks with GPT-5.4-mini and
  GPT-OSS-120B.
  The vertical dotted line marks the identifiability threshold $k^* = 1$
  (full source split).
  Bias is zero and omitted for \textit{A/P-Comp}.}
  \label{fig:coverage_sweep}
\end{figure*}
\subsection{Models}
\label{sec:conditions}

We evaluate a set of strong frontier proprietary and open models, including Qwen3.5-27B, Olmo3.1-32B-Instruct, GPT-OSS-120B, Llama-4-Maverick-17B-128E-Instruct, Gemini-3.1-Flash, and GPT-5.4-mini on all eight task types.

\noindent\textbf{Human study.}~To calibrate task difficulty, we recruited 7 participants (2 linguists, 5 crowdworkers via Upwork at an hourly rate of \$25) to attempt a subset of tasks (i.e., one variant per task type) under the same setups (see Appx.~\ref{appx:interface}).

\subsection{Evaluation Metrics}

We assess inductive reasoning through two complementary lenses: a \emph{direct} probe (can the model articulate the rule from demonstrations?) and an \emph{indirect}, behavioral probe (does the model achieve the tasks with high efficiency?).

\subsubsection{Rule Verbalization Score}
Given the source demonstrations, the model is asked to describe the hidden rule in its own words.
We use Qwen3.5-27B as a rubric-based LLM judge (see prompt in Appx.~\ref{appx:rule-verbalization-prompts}) to score this explanation against the ground-truth (Appx.~\ref{appx:llm-judge-ground-truth}) rule on a 0--2 scale, where 0 indicates an incorrect rule, 1 indicates partial recognition of the rule structure, and 2 indicates correct identification of the rule.

\noindent\textbf{LLM judge validation.}~Two human annotators (via Upwork, \$20/hr) independently rated a stratified sample of 99 items (33 per score level) using the same 0--2 rubric.
Inter-rater agreement was Krippendorff's $\alpha = 0.795$ (ordinal); human--LLM agreement was $\alpha = 0.84$ and $\alpha = 0.79$, matching human--human agreement, validating our judge.

\subsubsection{Efficiency-Calibrated Success Rate}
A model with strong inductive ability should not only solve gen tasks but do so efficiently. We capture this with the \emph{efficiency-calibrated success rate} (ECSR), defined as \textbf{$\text{success rate} \times \text{norm\_eff}$}.

\noindent\textbf{Task success rate.}~We report the \emph{task success rate}: the fraction of gen tasks in which the agent reaches the terminal goal within the episode budget.


\noindent\textbf{Norm\_eff.}~Task success alone does not distinguish a model that induces the rule from one that succeeds by enumeration. We therefore report \emph{normalized efficiency} for successful episodes. Since $n_\text{tries}$ varies across tasks, we normalize attempt count $t \in [1, n_\text{tries}]$ relative to the brute-force ceiling:
\[
  \text{norm\_eff} = \frac{n_\text{tries}/t - 1}{n_\text{tries} - 1} \in [0,1],
\]
where 0 corresponds to brute-force enumeration ($t {=} n_\text{tries}$) and 1 to solving in a single attempt ($t {=} 1$). 



\subsection{Results}
\label{decoupling}
We report mean ECSR and RV across all variants (Figure \ref{fig_ee_rv_combined}; Appx.~\ref{appx:full_results} Table \ref{tab:main_results}). Humans consistently outperform all models, with margins of 13\% on ECSR and 30\% on RV; the strongest models (GPT family) approach human-level ECSR on select attribute tasks but remain below human RV. Models perform better on attribute tasks than procedural ones, while humans maintain more balanced performance across both families.

Human data points show a clean bimodal pattern: high ECSR when RV is high, and lower ECSR when the rule is not identified, reflecting that humans who miss the rule generally cannot complete the task efficiently. Notably, even high RV does not always yield optimal ECSR among humans; we discuss this in Appx.~\ref{appx:full_results}.

\section{Do Models Act on the Rules?}
\label{sec:rule_induction}

In this section, we show that models show evidence of rule induction (\S\ref{sec:rv_ecsr_correlation}-\ref{sec:coverage_sweep}), but this capacity is uneven across tasks and limited. Alternatively, models may rely on other mechanisms, such as simply retrieving seen examples, to achieve task success without rule induction (\S\ref{sec:complications}).

\subsection{RV Correlates with ECSR}
\label{sec:rv_ecsr_correlation}

Across all models and tasks ($n=48$ model$\times$task points), ECSR and
RV are positively correlated (Spearman $\rho=0.54$, $p<0.001$): models
that score higher on rule verbalization also tend to perform more efficiently on the tasks.
A stronger positive trend is observed for humans ($\rho=0.85$, $p<0.001$), and the relationship is especially consistent on attribute induction tasks ($\rho=0.72$, $p<0.001$).
GPT-5.4-mini exemplifies this pattern most clearly. We provide further convergent evidence through a controlled manipulation of context availability in \S\ref{sec:coverage_sweep}, and then discuss patterns that may complicate this picture in \S\ref{sec:complications}.

\subsection{ECSR Tracks Identifiability}
\label{sec:coverage_sweep}

We further conduct a interventional study to manipulate how much rule-relevant information the context provides, from incomplete and underdetermined to fully identifiable, and ask whether task performance tracks this manipulation.
The key intuition is identifiability: for each task, there is a minimum source coverage $k^*$ below which the test rule $f/\sigma$ cannot be reliably recovered (\S\ref{AS}). 
If model success is due to the induction of $f/\sigma$, its performance at a coverage below $k^*$ will be lower.

\noindent\textbf{Setup.}~We vary coverage $k \in \{0, \frac{1}{3}k^*, \frac{2}{3}k^*, k^*,$ $2k^*\}$; at $k{=}0$ the context contains only distractor trajectories. The $2k^*$ condition isolates the contribution of demonstration quantity and signal strength, which may independently affect performance. We use GPT-5.4-mini and GPT-OSS-120B (both show non-trivial RV;\footnote{Non-trivial means average RV above 0.5, i.e., partial identification of the rule on average. Models with near-zero average RV demonstrate incorrect rule understanding and are poor candidates for probing rule-learning behaviors.} \S\ref{decoupling}). 


\noindent\textbf{Results.}~Figure~\ref{fig:coverage_sweep} shows ECSR across different coverages $k/k^*$ for all tasks.
The dominant pattern across most tasks is a rise toward $k^*$, consistent with task performance improving as demonstrations become sufficient to recover the rule, followed by a plateau at $2k^*$: performance does not continue to improve once the identifiability threshold is reached, suggesting the gains are attributable to rule identifiability rather than to demonstration quantity or signal strength alone (\textit{A-Comp}, \textit{A-Over}, \textit{P-Add}, and \textit{P-Over}).
Some tasks, e.g., \textit{A-Cond}, show a striking reversal that we attribute to a different mechanism, discussed in \S\ref{sec:complications}.


\subsection{Models Are Still Poor Reasoners}
\label{sec:complications}
Despite the evidence above, models are still able to succeed with a low RV performance; we show that this is consistent with copying (from the context).

\noindent\textbf{Where RV and ECSR dissociate.}~
While the overall RV:ECSR correlation is positive, it is much weaker for procedural tasks ($\rho=0.36$, $p=0.08$). More strikingly, a cluster of model$\times$task points accumulates at near-zero RV while spanning a wide range of ECSR values (Figure~\ref{fig_ee_rv_combined}): even at near-zero RV, models achieve ECSR that exceeds what humans achieve under similarly low RV. This is most visible for Gemini.
Such non-trivial performance but with low RV echoes prior claims that LLMs are still poor rule-based reasoners \citep{li2025mirage}.


\noindent\textbf{Contextual bias.}~
For some tasks (e.g., \textit{A-Cond}), ECSR spikes at intermediate $k$ and then declines: models do not faithfully track the progressive reveal of rule-relevant context.
We attribute this to \emph{contextual bias}: at low $k$, the source demonstrations may include the target item/process directly, and with a small option space the model may simply copy a seen item/process rather than induce the rule.
We measure contextual bias as the expected success rate of a model guessing uniformly among seen items:
\[
  \text{contextual\_bias}(k) = \mathbb{E}\!\left[
    \frac{\mathbf{1}[d^{*} \in \mathcal{D}_k]}{|\mathcal{D}_k|}
  \right],
\]
where $\mathcal{D}_k$ is the set of distinct items/processes in the $k$ source demonstrations.
As shown in Figure \ref{fig:coverage_sweep}, for \textit{A-Cond}, ECSR actually tracks the bias curve rather than coverage of rule context, indicating that performance here is most likely to be driven by copying rather than rule induction.


\section{Multi-Step Execution Adds Complexity}
\label{sec:other_factors}

One feature of \AS is that rule application is embedded in a multi-step execution setting, reflecting how agents deploy inductive knowledge in practice. We ask whether this grounding adds difficulty beyond identifying the rule itself: we compare \emph{direct-answer} (QA), where the model states the correct item or process in a single turn, against \emph{episodic} execution, where it must act out the solution across multiple turns. Experiments use GPT-5.4-mini and Qwen3.5-27B.


\begin{table}[t]
\centering
\small
\caption{Format gap ($\Delta\mu = \text{QA} - \text{ECSR}$) aggregated by model and task family.
\colorbox{orange!25}{Orange}: execution bottleneck (QA $>$ ECSR);
\colorbox{blue!15}{blue}: QA bottleneck (ECSR $>$ QA).}
\label{tab:format_gap}
\begin{tabular*}{\columnwidth}{@{\extracolsep{\fill}} l r r}
\toprule
\textbf{Models} & \textbf{Attr.} & \textbf{Proc.} \\
\midrule
GPT-5.4-mini        & \cellcolor{orange!25}$\phantom{-}0.096$\textsuperscript{**\phantom{*}}  & \cellcolor{orange!25}$\phantom{-}0.268$\textsuperscript{***} \\
Qwen3.5-27B  & \cellcolor{orange!25}$\phantom{-}0.196$\textsuperscript{***} & \cellcolor{blue!15}$-0.111$\textsuperscript{*\phantom{**}} \\
\bottomrule
\end{tabular*}
\end{table}



We define the \emph{format gap} as \textbf{QA accuracy minus ECSR}. A positive gap (QA $>$ ECSR) indicates an \emph{execution bottleneck}: behavioral success in the QA condition does not transfer to process execution. A negative gap (ECSR $>$ QA) indicates a \emph{QA bottleneck}: the tasks are inherently hard for these models, while exploration within the environment helps the performance. A gap near zero indicates that the format does not introduce extra complexity. We test whether the per-variant gap ($N{=}160$; 20 variants $\times$ 8 tasks) differs from zero via sign-permutation test ($5{,}000$ permutations, Bonferroni-corrected by task family; $^{**}p{<}0.01$, $^{***}p{<}0.001$).

\noindent\textbf{Results.}~Both models show a positive gap on attribute tasks (Table~\ref{tab:format_gap}; per-task in Appx.~\ref{appx:full_results}): correctly identifying the rule does not guarantee successful episodic execution. For GPT-5.4-mini this extends to procedural tasks ($\Delta\mu{=}0.268$, $p{<}0.001$). Qwen3.5-27B reverses on procedural tasks ($\Delta\mu{=}{-}0.11$, $p{=}0.010$), where episodic execution outpaces direct answering, suggesting procedural tasks are inherently harder for this model.

We further examine one additional factor shaping performance: surface semantics (nonce vs.\ semantic names). The effect is minimal overall: GPT-5.4-mini is insensitive to lexical condition, and Qwen3.5-27B shows only a small advantage for nonce names on attribute tasks (Appx.~\ref{appx:nonce}). We also explore whether existing induction methods such as ReAct \citep{yao2023react}, ACE \citep{zhang2026agentic}, HR \citep{qiu-etal-2024-phenomenal}, and IDEA \citep{he2024idea}, improve performance, finding limited gains concentrated on attribute tasks with no reliable improvement on procedural tasks; see Appx.~\ref{appx:methods}.

\section{Conclusion}


 Having established a principled benchmark for rule induction in goal-directed episodic tasks,  we believe \AS represents a step toward a new class of evaluation for the inductive reasoning ability of AI agents: in addition to asking what a model knows, it asks  
  whether a model can learn a rule from experience and act correctly on it in a novel situation, an important ability as LLMs are increasingly deployed in settings, where adapting to unspecified rules is essential. We hope our design provides a configurable testbed for future work, including how inductive capacity scales with rule interaction complexity, how demonstration  quantity and diversity shape induction beyond the identifiability minimum, and what architectural and prompting choices close the gap between the ability to induce the rules and
  executing them.

\section*{Limitations}
\label{sec:limitations}

While \AS is designed to be customizable, extending it beyond the provided task types requires manual effort: new rule interactions and their identifiability conditions must be worked out by the practitioner if the current conditions do not apply.
\AS is fully synthetic; how the observed failures transfer to naturalistic settings with noisy evidence and open-ended action spaces remains open, though the framework is also ready to accommodate such extensions.
Our metrics do not fully rule out trial-and-error as a path to success: while rule induction tends to improve efficiency, and high ECSR could serve as a proxy of good rule induction, it is not always the case, and we leave the development of more direct causal indicators to future work. The RV judge (Qwen3.5-27B) shares a model family with one of the evaluated systems; notably, Qwen achieves among the lowest RV scores in our results, suggesting no systematic inflation, but future work is encouraged to corroborate with a disjoint-family or ensemble judge for additional confidence.
Our human baseline is drawn from a small participant pool and should be treated as indicative rather than definitive.
Finally, the methods evaluation is intended as an initial exploration; training-based approaches and other paradigms remain to be investigated.

\section*{Acknowledgments}
We thank William Sheffield and other annotators for their annotations as well as Elias Stengel-Eskin and UT Computational Linguistics lab members for their feedback. 
This work was partially supported by NSF grants IIS-2145479 and IIS-2107524, and Good Systems,\footnote{https://goodsystems.utexas.edu/} a UT Austin Grand
Challenge to develop responsible AI technologies.
We thank the Texas Advanced Computing Center
(TACC)\footnote{http://www.tacc.utexas.edu/} at UT Austin for providing computational
resources that have contributed to the research results reported within this paper. KM acknowledges support from the Donald D. Harrington Faculty Fellowship at UT Austin.

\bibliography{custom}
\clearpage
\appendix

\section{Notation for \S\ref{attribute induction}--\ref{procedural induction}}
\label{appx:notation}

\begin{tcolorbox}[
  colback=gray!4, colframe=black!20, boxrule=0.5pt, arc=2pt,
  left=6pt, right=6pt, top=4pt, bottom=4pt
]
\small
\begin{tabular}{@{}l@{\quad}p{0.62\columnwidth}@{}}
\toprule
\textbf{Symbol} & \textbf{Description} \\
\midrule
$v_1$ & class attribute value \\
$v_2$ & role attribute value \\
$f(v_1,v_2)$ & hidden item mapping (attribute induction) \\
$\sigma(v_1,v_2)$ & hidden process selector (procedural induction) \\
$R(v_1)$ & regime function ($\in\{0,1\}$, partitions $V_1$ into two groups) \\
$u^*$ & designated override role value \\
$a$ & action type of inserted node \\
$o$ & ordering of inserted node relative to \texttt{buy} \\
$\text{size}(\cdot),\text{color}(\cdot)$ & item dimension functions (attribute value $\to$ size or color) \\
$\text{count}(\cdot)$ & node count function (Proc1: attribute value $\to$ integer count) \\
$\text{act}(\cdot),\text{ord}(\cdot)$ & process knob functions (attribute value $\to$ $a$ or $o$) \\
$\text{size}_0,\text{color}_0$ & fixed per-regime constants (Attr3) \\
$\text{act}_0,\text{ord}_0$ & fixed per-regime constants (Proc3) \\
$a_0, o_0$ & fixed override knobs (Proc4) \\
\bottomrule
\end{tabular}
\end{tcolorbox}

\section{Task Instantiation Details}
\label{appendix:task_details}

\begin{figure*}[htb]
  \centering
  \includegraphics[width=\textwidth]{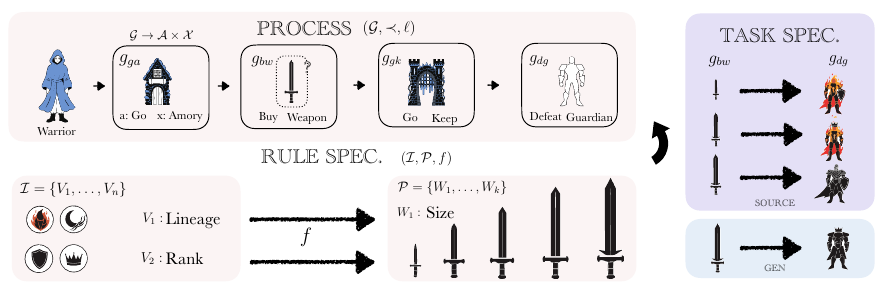}
  \caption{Task curation illustrated on the Dragon's Keep example (cf.\ Figure~\ref{fig:task_overview}).
  \textit{Top left} (\textbf{Process}): goal network $(\mathcal{G}, \prec, \ell)$.
  \textit{Bottom left} (\textbf{Rule Specification}): independent dimensions $\mathcal{I}$ mapped via $f$ to dependent dimension $\mathcal{P}$.
  \textit{Right} (\textbf{Task Specification}): goal nodes bound to their dimensions; source and gen splits shown.}
  \label{fig:task_curation}
\end{figure*}

\tikzset{
  sc/.style={draw=gray!55, fill=gray!22, minimum width=1.0cm, minimum height=0.90cm,
             inner sep=2pt, font=\scriptsize, align=center},
  gc/.style={draw=black, line width=0.9pt, fill=white, minimum width=1.0cm, minimum height=0.90cm,
             inner sep=2pt, font=\scriptsize\bfseries, align=center},
  scc/.style={draw=gray!55, fill=blue!18, minimum width=1.0cm, minimum height=0.90cm,
              inner sep=2pt, font=\scriptsize, align=center},
  gcc/.style={draw=black, line width=0.9pt, fill=blue!18, minimum width=1.0cm, minimum height=0.90cm,
              inner sep=2pt, font=\scriptsize\bfseries, align=center},
  scs/.style={draw=gray!55, fill=green!20, minimum width=1.0cm, minimum height=0.90cm,
              inner sep=2pt, font=\scriptsize, align=center},
  gcs/.style={draw=black, line width=0.9pt, fill=green!20, minimum width=1.0cm, minimum height=0.90cm,
              inner sep=2pt, font=\scriptsize\bfseries, align=center},
  ov/.style={draw=red!55, fill=red!14, minimum width=1.0cm, minimum height=0.90cm,
             inner sep=2pt, font=\scriptsize, align=center},
  gov/.style={draw=black, line width=0.9pt, fill=red!14, minimum width=1.0cm, minimum height=0.90cm,
              inner sep=2pt, font=\scriptsize\bfseries, align=center},
  hd/.style={font=\scriptsize\bfseries}
}

\subsection{Task Generation Procedure}
\label{appx:task_generation}

A task specification binds process template(s) $\Pi$, a rule specification $(\mathcal{I}, \mathcal{P}, f)$, and a lexicon $\mathcal{L}$. It declares (i) a set of process templates $\{\Pi_1,\ldots,\Pi_m\}$, (ii) a rule specification that specifies what rule applies in the task, and (iii) for each $\Pi_i$, which subgoal nodes carry which attributes. Task generation proceeds in three steps:

\begin{enumerate}
  \item \textbf{Rule binding.} Bind the independent attributes to the designated nodes of $\Pi_i$. If the task includes a mapping $f/\sigma$, resolve the dependent variables by binding item properties or adding new sub-goal nodes. Tasks may include both $\sigma$ and $f$.
  \item \textbf{Lexicon instantiation.} Replace all abstract attribute values, entity names, item names, and locations with surface forms from $\mathcal{L}$: \emph{semantic} mode uses naturalistic names (e.g., \emph{dragon-blooded}, \emph{obsidian blade}); \emph{nonce} mode uses pronounceable nonsense syllables (e.g., \emph{krev\_A}, \emph{vrel\_Z}), stripping any advantage from world knowledge.
\end{enumerate}

In the Dragon's Keep example, a uniform process is used, \{lineage, rank\} are bound to the guardian node and \{size\} to the weapon node via $f$. The agent must induce the size of the required weapon from demonstrations.

\subsection{Identifiability Conditions}
\label{appx:identifiability}

For each task, we construct source demonstrations so that the target rule is recoverable from the demonstrations rather than leaving structurally distinct rules equally consistent with the data.
The conditions are shared across the attribute and procedural families, which have the same composition structures (applied to $f$ and $\sigma$ respectively).
For each rule type we describe the rule, explain why the condition is needed, and illustrate the ambiguity that arises without it.

\noindent\textbf{Additive (A-Add / P-Add).}~The rule combines two independent scalar mappings: $f(v_1,v_2)=\phi_1(v_1)+\phi_2(v_2)$. Because only the sum is observed, recovering each mapping requires contrasting observations that share one attribute value while the other varies. The source pairs must therefore form a \emph{connected} bipartite graph over $V_1\times V_2$.

\smallskip
\noindent\textit{Counterexample}: suppose source contains only two observations:
\begin{center}\footnotesize
\begin{tabular}{l|ccc}
 & \textit{prophet} & \textit{berserker} & \textit{chirurgeon} \\\hline
\textit{ranger}   & 4 & \textemdash & \textemdash \\
\textit{merchant} & \textemdash & 1 & \textemdash \\
\textit{captain}  & \textemdash & \textemdash & \textemdash \\
\end{tabular}
\end{center}
The two components are disconnected: $\phi_1(\text{ranger}){+}\phi_2(\text{berserker})$ is entirely unconstrained, so multiple additive rules are equally consistent with the data.

\medskip
\noindent\textbf{Compositional (A-Comp / P-Comp).}~The rule maps each attribute to a different output dimension independently: $f(v_1,v_2)=\bigl(s(v_1),\,c(v_2)\bigr)$. Because the model has no prior knowledge that $s$ and $c$ are independent, independence must be \emph{observable} from demonstrations---requiring pairs where one attribute varies while the other is fixed. This again requires a connected bipartite source graph.

\smallskip
\noindent\textit{Counterexample}: suppose source contains only:
\begin{center}\footnotesize
\begin{tabular}{l|ccc}
 & \textit{prophet} & \textit{berserker} & \textit{chirurgeon} \\\hline
\textit{ranger}  & (large, red) & \textemdash & \textemdash \\
\textit{captain} & \textemdash & (short, grey) & \textemdash \\
\end{tabular}
\end{center}
With no shared partners, the contribution of \textit{ranger} is ambiguous: it could map to \textit{large}, to \textit{red}, or to nothing if \textit{prophet} alone determines both dimensions. A conjunctive lookup table is equally consistent.

\medskip
\noindent\textbf{Conditional (A-Cond / P-Cond).}~The rule assigns $v_1$ to a regime via $R:V_1\to\{0,1\}$, then uses $v_2$ to control a dimension specific to that regime. Two conditions are required.

\emph{(1) Regime identification}: each $V_1$ value must appear with at least two distinct $V_2$ values. Without it, if \textit{ranger} is observed only with \textit{berserker}$\to$(\textit{large}, \textit{crimson}), both ``ranger is in the size regime'' and ``ranger is in the color regime'' are consistent with that single observation---a second $v_2$ partner is needed to reveal which dimension varies with role.

\emph{(2) Within-regime coverage}: each $V_2$ value must appear within each regime. Without it, if \textit{mender} never co-occurs with a regime-1 class, its within-regime mapping is never observed, leaving a gen entity in regime~1 with role \textit{mender} unconstrained.

\medskip
\noindent\textbf{Override (A-Over / P-Over).}~The rule applies a base mapping $s:V_1\to I$ unless $v_2=u^*$, in which case a fixed override item $h$ is always selected. Two \emph{source identifiability} conditions apply: \emph{(i)} every $V_1$ value must appear with at least one non-override $v_2$, so $s(v_1)$ is observed for each class; \emph{(ii)} $u^*$ must appear with at least two distinct $V_1$ values---otherwise, if only (\textit{ranger}, $u^*$)$\to h$ is seen, ``$u^*$ always triggers $h$'' is indistinguishable from ``ranger always uses $h$''. Two \emph{gen-split design} conditions ensure the test is non-trivial: \emph{(iii)} one class is withheld from override observations in source, so the gen split tests extrapolation of the override to a novel (class, $u^*$) pair; \emph{(iv)} the gen split also includes a non-override pair for that class, confirming the base rule applies otherwise.

\smallskip
\noindent\textit{Illustration of the gen split design} (captain withheld from source override):
\begin{center}\footnotesize
\begin{tabular}{l|cc}
 & $u^*$ \textit{(chirurgeon)} & \textit{other roles} \\\hline
\textit{ranger}   & $h$ \textnormal{(source)} & $s(\text{ranger})$ \textnormal{(source)} \\
\textit{merchant} & $h$ \textnormal{(source)} & $s(\text{merchant})$ \textnormal{(source)} \\
\textit{captain}  & $h$ \textnormal{(gen, iii)} & $s(\text{captain})$ \textnormal{(gen, iv)} \\
\end{tabular}
\end{center}

\begin{figure*}[htb]
\centering
\small

\begin{minipage}[t]{0.47\textwidth}
\centering
{\small\textbf{(a) A-Add: Additive}\quad $f=\phi_1(v_1)+\phi_2(v_2)$}\\[5pt]
\begin{tikzpicture}
  \node[hd,rotate=45,anchor=south west] at (0.27,0.05) {prophet};
  \node[hd,rotate=45,anchor=south west] at (1.37,0.05) {berserker};
  \node[hd,rotate=45,anchor=south west] at (2.47,0.05) {chirurgeon};
  \node[hd,anchor=east] at (-0.10,-0.50) {ranger};
  \node[hd,anchor=east] at (-0.10,-1.50) {merchant};
  \node[hd,anchor=east] at (-0.10,-2.50) {captain};
  \node[sc] at (0.55,-0.50) {4}; \node[sc] at (1.65,-0.50) {3}; \node[gc] at (2.75,-0.50) {2};
  \node[gc] at (0.55,-1.50) {3}; \node[sc] at (1.65,-1.50) {2}; \node[sc] at (2.75,-1.50) {1};
  \node[sc] at (0.55,-2.50) {2}; \node[gc] at (1.65,-2.50) {1}; \node[sc] at (2.75,-2.50) {0};
\end{tikzpicture}\\[4pt]
{\scriptsize cell = item size\quad rows: class\quad cols: role}
\end{minipage}
\hfill
\begin{minipage}[t]{0.47\textwidth}
\centering
{\small\textbf{(b) A-Comp: Compositional}\quad $f=\bigl(s(v_1),\,c(v_2)\bigr)$}\\[5pt]
\begin{tikzpicture}
  \node[hd,rotate=45,anchor=south west] at (0.27,0.05) {prophet};
  \node[hd,rotate=45,anchor=south west] at (1.37,0.05) {berserker};
  \node[hd,rotate=45,anchor=south west] at (2.47,0.05) {chirurgeon};
  \node[hd,anchor=east] at (-0.10,-0.50) {ranger};
  \node[hd,anchor=east] at (-0.10,-1.50) {merchant};
  \node[hd,anchor=east] at (-0.10,-2.50) {captain};
  \node[sc] at (0.55,-0.50) {\shortstack{colossal\\crimson}};
  \node[sc] at (1.65,-0.50) {\shortstack{colossal\\grey}};
  \node[gc] at (2.75,-0.50) {\shortstack{colossal\\purple}};
  \node[gc] at (0.55,-1.50) {\shortstack{long\\crimson}};
  \node[sc] at (1.65,-1.50) {\shortstack{long\\grey}};
  \node[sc] at (2.75,-1.50) {\shortstack{long\\purple}};
  \node[sc] at (0.55,-2.50) {\shortstack{standard\\crimson}};
  \node[gc] at (1.65,-2.50) {\shortstack{standard\\grey}};
  \node[sc] at (2.75,-2.50) {\shortstack{standard\\purple}};
\end{tikzpicture}\\[4pt]
{\scriptsize cell = (size, color) item}
\end{minipage}

\vspace{10pt}

\begin{minipage}[t]{0.47\textwidth}
\centering
{\small\textbf{(c) A-Cond: Conditional}\quad $v_1\!\to\!\text{regime}$; regime sets $v_2$'s dim}\\[5pt]
\begin{tikzpicture}
  \node[hd,rotate=45,anchor=south west] at (0.27,0.05) {berserker};
  \node[hd,rotate=45,anchor=south west] at (1.37,0.05) {chirurgeon};
  \node[hd,rotate=45,anchor=south west] at (2.47,0.05) {mender};
  \node[hd,anchor=east] at (-0.10,-0.50) {ranger};
  \node[hd,anchor=east] at (-0.10,-1.50) {merchant};
  \node[hd,anchor=east] at (-0.10,-2.50) {captain};
  \node[hd,anchor=east] at (-0.10,-3.50) {minstrel};
  \node[font=\tiny,text=green!60!black,anchor=west] at (3.38,-1.00) {\textit{size regime}};
  \node[font=\tiny,text=blue!65,anchor=west]        at (3.38,-3.00) {\textit{color regime}};
  \draw[dashed,gray!50,thin] (-1.00,-2.00) -- (3.38,-2.00);
  \node[scs] at (0.55,-0.50) {\shortstack{crimson\\short}};
  \node[scs] at (1.65,-0.50) {\shortstack{crimson\\great}};
  \node[gcs] at (2.75,-0.50) {\shortstack{crimson\\colossal}};
  \node[gcs] at (0.55,-1.50) {\shortstack{crimson\\short}};
  \node[scs] at (1.65,-1.50) {\shortstack{crimson\\great}};
  \node[scs] at (2.75,-1.50) {\shortstack{crimson\\colossal}};
  \node[scc] at (0.55,-2.50) {\shortstack{crimson\\long}};
  \node[scc] at (1.65,-2.50) {\shortstack{silver\\long}};
  \node[gcc] at (2.75,-2.50) {\shortstack{white\\long}};
  \node[gcc] at (0.55,-3.50) {\shortstack{crimson\\long}};
  \node[scc] at (1.65,-3.50) {\shortstack{silver\\long}};
  \node[scc] at (2.75,-3.50) {\shortstack{white\\long}};
\end{tikzpicture}\\[4pt]
{\scriptsize cell = (color, size) item\quad green=size-varies regime\quad blue=color-varies regime}
\end{minipage}
\hfill
\begin{minipage}[t]{0.47\textwidth}
\centering
{\small\textbf{(d) A-Over: Override}\quad $u^*{=}$chirurgeon, $h{=}$great}\\[5pt]
\begin{tikzpicture}
  \node[hd,rotate=45,anchor=south west] at (0.27,0.05) {prophet};
  \node[hd,rotate=45,anchor=south west] at (1.37,0.05) {berserker};
  \node[hd,rotate=45,anchor=south west,text=red!65] at (2.47,0.05) {chirurgeon$^*$};
  \node[hd,anchor=east] at (-0.10,-0.50) {ranger};
  \node[hd,anchor=east] at (-0.10,-1.50) {merchant};
  \node[hd,anchor=east] at (-0.10,-2.50) {captain};
  \node[sc]  at (0.55,-0.50) {standard}; \node[sc]  at (1.65,-0.50) {standard}; \node[ov]  at (2.75,-0.50) {great};
  \node[gc]  at (0.55,-1.50) {long};     \node[sc]  at (1.65,-1.50) {long};     \node[gov] at (2.75,-1.50) {great};
  \node[sc]  at (0.55,-2.50) {colossal}; \node[sc]  at (1.65,-2.50) {colossal}; \node[ov]  at (2.75,-2.50) {great};
\end{tikzpicture}\\[4pt]
{\scriptsize cell = item size\quad merchant=held-out class\quad $^*$=override role}
\end{minipage}

\vspace{6pt}
{\scriptsize
\colorbox{gray!22}{\phantom{i}}~source\quad
\fbox{\phantom{i}}~gen target\quad
\colorbox{green!20}{\phantom{i}}~size-varies regime\quad
\colorbox{blue!18}{\phantom{i}}~color-varies regime\quad
\colorbox{red!14}{\phantom{i}}~override
}

\caption{Attribute induction task grids (one instantiation of the source/gen split).
Rows index class ($v_1 \in V_1$), columns index role ($v_2 \in V_2$).
Shaded cells are source demonstrations; \textbf{bold-outlined} cells are held-out gen targets.
The split shown is a single representative draw; other valid splits satisfying the identifiability conditions are possible.}
\label{fig:prop_grids}
\end{figure*}
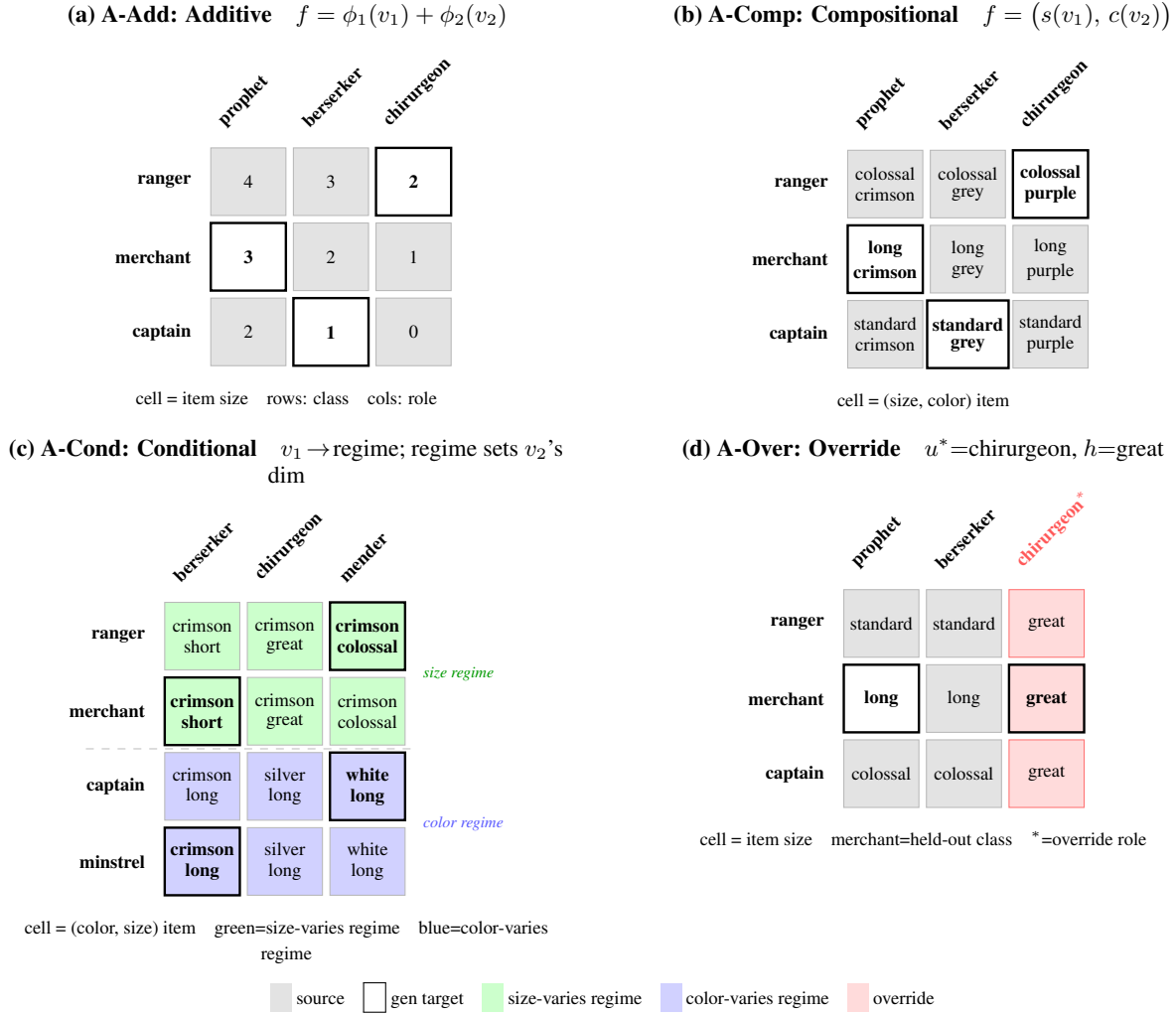

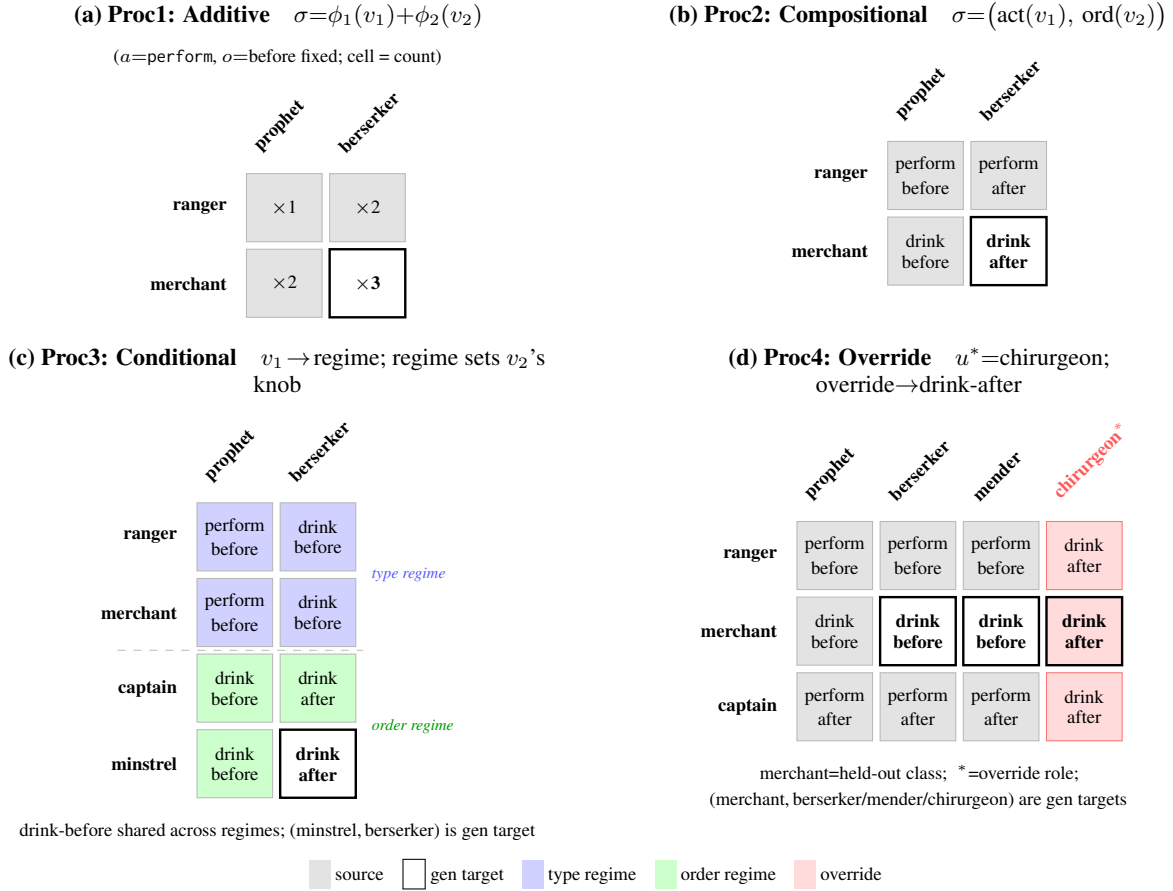
\begin{figure*}[htb]
\centering
\small

\begin{minipage}[t]{0.47\textwidth}
\centering
{\small\textbf{(a) Proc1: Additive}\quad $\sigma{=}\phi_1(v_1){+}\phi_2(v_2)$}\\[5pt]
{\scriptsize ($a{=}$\texttt{perform}, $o{=}$before fixed; cell = count)}\\[4pt]
\begin{tikzpicture}
  \node[hd,rotate=45,anchor=south west] at (0.27,0.05) {prophet};
  \node[hd,rotate=45,anchor=south west] at (1.37,0.05) {berserker};
  \node[hd,anchor=east] at (-0.10,-0.50) {ranger};
  \node[hd,anchor=east] at (-0.10,-1.50) {merchant};
  \node[sc] at (0.55,-0.50) {$\times$1}; \node[sc] at (1.65,-0.50) {$\times$2};
  \node[sc] at (0.55,-1.50) {$\times$2}; \node[gc] at (1.65,-1.50) {$\times$3};
\end{tikzpicture}
\end{minipage}
\hfill
\begin{minipage}[t]{0.47\textwidth}
\centering
{\small\textbf{(b) Proc2: Compositional}\quad $\sigma{=}\bigl(\text{act}(v_1),\,\text{ord}(v_2)\bigr)$}\\[5pt]
\begin{tikzpicture}
  \node[hd,rotate=45,anchor=south west] at (0.27,0.05) {prophet};
  \node[hd,rotate=45,anchor=south west] at (1.37,0.05) {berserker};
  \node[hd,anchor=east] at (-0.10,-0.50) {ranger};
  \node[hd,anchor=east] at (-0.10,-1.50) {merchant};
  \node[sc] at (0.55,-0.50) {\shortstack{perform\\before}};
  \node[sc] at (1.65,-0.50) {\shortstack{perform\\after}};
  \node[sc] at (0.55,-1.50) {\shortstack{drink\\before}};
  \node[gc] at (1.65,-1.50) {\shortstack{drink\\after}};
\end{tikzpicture}
\end{minipage}

\vspace{10pt}

\begin{minipage}[t]{0.47\textwidth}
\centering
{\small\textbf{(c) Proc3: Conditional}\quad $v_1\!\to\!\text{regime}$; regime sets $v_2$'s knob}\\[5pt]
\begin{tikzpicture}
  \node[hd,rotate=45,anchor=south west] at (0.27,0.05) {prophet};
  \node[hd,rotate=45,anchor=south west] at (1.37,0.05) {berserker};
  \node[hd,anchor=east] at (-0.10,-0.50) {ranger};
  \node[hd,anchor=east] at (-0.10,-1.50) {merchant};
  \node[hd,anchor=east] at (-0.10,-2.50) {captain};
  \node[hd,anchor=east] at (-0.10,-3.50) {minstrel};
  \node[font=\tiny,text=blue!65,anchor=west] at (2.25,-1.00) {\textit{type regime}};
  \node[font=\tiny,text=green!60!black,anchor=west] at (2.25,-3.00) {\textit{order regime}};
  \draw[dashed,gray!50,thin] (-1.00,-2.00) -- (2.25,-2.00);
  \node[scc] at (0.55,-0.50) {\shortstack{perform\\before}};
  \node[scc] at (1.65,-0.50) {\shortstack{drink\\before}};
  \node[scc] at (0.55,-1.50) {\shortstack{perform\\before}};
  \node[scc] at (1.65,-1.50) {\shortstack{drink\\before}};
  \node[scs] at (0.55,-2.50) {\shortstack{drink\\before}};
  \node[scs] at (1.65,-2.50) {\shortstack{drink\\after}};
  \node[scs] at (0.55,-3.50) {\shortstack{drink\\before}};
  \node[gc]  at (1.65,-3.50) {\shortstack{drink\\after}};
\end{tikzpicture}\\[4pt]
{\scriptsize drink-before shared across regimes; (minstrel,\,berserker) is gen target}
\end{minipage}
\hfill
\begin{minipage}[t]{0.47\textwidth}
\centering
{\small\textbf{(d) Proc4: Override}\quad $u^*{=}$chirurgeon; override$\to$drink-after}\\[5pt]
\begin{tikzpicture}
  \node[hd,rotate=45,anchor=south west] at (0.27,0.05) {prophet};
  \node[hd,rotate=45,anchor=south west] at (1.37,0.05) {berserker};
  \node[hd,rotate=45,anchor=south west] at (2.47,0.05) {mender};
  \node[hd,rotate=45,anchor=south west,text=red!65] at (3.57,0.05) {chirurgeon$^*$};
  \node[hd,anchor=east] at (-0.10,-0.50) {ranger};
  \node[hd,anchor=east] at (-0.10,-1.50) {merchant};
  \node[hd,anchor=east] at (-0.10,-2.50) {captain};
  \node[sc]  at (0.55,-0.50) {\shortstack{perform\\before}};
  \node[sc]  at (1.65,-0.50) {\shortstack{perform\\before}};
  \node[sc]  at (2.75,-0.50) {\shortstack{perform\\before}};
  \node[ov]  at (3.85,-0.50) {\shortstack{drink\\after}};
  \node[sc]  at (0.55,-1.50) {\shortstack{drink\\before}};
  \node[gc]  at (1.65,-1.50) {\shortstack{drink\\before}};
  \node[gc]  at (2.75,-1.50) {\shortstack{drink\\before}};
  \node[gov] at (3.85,-1.50) {\shortstack{drink\\after}};
  \node[sc]  at (0.55,-2.50) {\shortstack{perform\\after}};
  \node[sc]  at (1.65,-2.50) {\shortstack{perform\\after}};
  \node[sc]  at (2.75,-2.50) {\shortstack{perform\\after}};
  \node[ov]  at (3.85,-2.50) {\shortstack{drink\\after}};
\end{tikzpicture}\\[4pt]
{\scriptsize merchant=held-out class;\; $^*$=override role;\; (merchant,\,berserker/mender/chirurgeon) are gen targets}
\end{minipage}

\vspace{6pt}
{\scriptsize
\colorbox{gray!22}{\phantom{i}}~source\quad
\fbox{\phantom{i}}~gen target\quad
\colorbox{blue!18}{\phantom{i}}~type regime\quad
\colorbox{green!20}{\phantom{i}}~order regime\quad
\colorbox{red!14}{\phantom{i}}~override
}

\caption{Procedural induction task grids (one instantiation of the source/gen split).
Rows index class ($v_1 \in V_1$), columns index role ($v_2 \in V_2$).
Shaded cells are source demonstrations; \textbf{bold-outlined} cells are held-out gen targets.
Each cell shows the process template $\sigma(v_1,v_2)$ (action type and ordering).
The split shown is a single representative draw; other valid splits satisfying the identifiability conditions are possible.}
\label{fig:proc_grids}
\end{figure*}

Tables~\ref{tab:prop_tasks} and~\ref{tab:proc_tasks} summarize
cardinalities and split sizes; Figures~\ref{fig:prop_grids}
and~\ref{fig:proc_grids} show the concrete source/gen splits as grids.

\subsection{Attribute Induction: Concrete Instantiations}

All attribute induction tasks use $n=2$ independent attribute dimensions bound to the entity node: \texttt{class}~($V_1$) and \texttt{role}~($V_2$).

\begin{sloppypar}
\paragraph{Attr1 (Additive)}
$|V_1|{=}|V_2|{=}3$: class $\in \{$ranger, merchant, captain$\}$; role $\in \{$prophet, berserker, chirurgeon$\}$.
$\phi_1$: ranger$\to 2$, merchant$\to 1$, captain$\to 0$.
$\phi_2$: prophet$\to 2$, berserker$\to 1$, chirurgeon$\to 0$.
Output: item sizes 0--4.

\paragraph{Attr2 (Compositional)}
Same attribute sets as \textit{A-Add}.
$s$: ranger$\to$colossal, merchant$\to$long, captain$\to$standard.
$c$: prophet$\to$crimson, berserker$\to$grey, chirurgeon$\to$purple.
Output: $3{\times}3{=}9$ (size, color) items.

\paragraph{Attr3 (Conditional)}
$|V_1|{=}4$: class $\in \{$ranger, merchant, captain, minstrel$\}$;
$|V_2|{=}3$: role $\in \{$berserker, chirurgeon, mender$\}$.
$R$: ranger, merchant $\to$ regime~0; captain, minstrel $\to$ regime~1.
Regime~0 (color fixed at crimson): $f_\text{size}$: berserker$\to$short, chirurgeon$\to$great, mender$\to$colossal.
Regime~1 (size fixed at long): $f_\text{color}$: berserker$\to$crimson, chirurgeon$\to$silver, mender$\to$white.
Output: $3{\times}4{=}12$ (color, size) Cartesian combinations; 6 are achievable under the regime structure, matching $n_\text{tries}{=}6$.

\paragraph{Attr4 (Override)}
Same attribute sets as \textit{A-Add}/\textit{A-Comp}.
$u^*{=}$chirurgeon, $h{=}$great (override size).
$s$: ranger$\to$standard, captain$\to$colossal; merchant$\to$long (held-out class in source).
Output: 4 sizes (standard, long, colossal, great).
\end{sloppypar}

\subsection{Procedural Induction: Concrete Instantiations}

All procedural induction tasks use the same attribute dimensions as attribute induction tasks: \texttt{class}~($V_1$) and \texttt{role}~($V_2$). The ordering set $O$ encodes precedence relations between the inserted node $g_\text{new}$ and the reference node $g_\text{ref}{=}\texttt{buy}$: $O{=}\{g_\text{new}\prec g_\text{ref},\;g_\text{ref}\prec g_\text{new}\}$, abbreviated \texttt{before} and \texttt{after} respectively.

\begin{sloppypar}
\paragraph{Proc1 (Additive)}
$|V_1|{=}|V_2|{=}2$: class $\in \{$ranger, merchant$\}$; role $\in \{$prophet, berserker$\}$.
$\phi_1$: ranger$\to 1$, merchant$\to 2$.
$\phi_2$: prophet$\to 0$, berserker$\to 1$.
Fixed: $a{=}$\texttt{perform}, $o{=}$\texttt{before}; count $\phi_1(v_1)+\phi_2(v_2)\in\{1,2,3\}$.

\paragraph{Proc2 (Compositional)}
Same attribute sets as Proc1.
$\text{act}$: ranger$\to$\texttt{perform}, merchant$\to$\texttt{drink}.
$\text{ord}$: prophet$\to$\texttt{before}, berserker$\to$\texttt{after}.
$|\Pi|{=}4$: perform-before, perform-after, drink-before, drink-after.

\paragraph{Proc3 (Conditional)}
$|V_1|{=}4$: class $\in \{$ranger, merchant, captain, minstrel$\}$;
$|V_2|{=}2$: role $\in \{$prophet, berserker$\}$.
$R$: ranger, merchant $\to$ regime~0; captain, minstrel $\to$ regime~1.
Regime~0 ($o{=}$before fixed): $a$: prophet$\to$\texttt{perform}, berserker$\to$\texttt{drink}.
Regime~1 ($a{=}$\texttt{drink} fixed): $o$: prophet$\to$\texttt{before}, berserker$\to$\texttt{after}.
$|\Pi|{=}3$: perform-before, drink-before (shared across regimes), drink-after.

\paragraph{Proc4 (Override)}
$|V_1|{=}3$: class $\in \{$ranger, merchant, captain$\}$ (merchant is the held-out class);
$|V_2|{=}4$: role $\in \{$prophet, berserker, mender, chirurgeon$\}$ with $u^*{=}$chirurgeon.
$\Pi_\mathrm{ov}{=}$drink-after (chirurgeon triggers override for any class).
$\Pi_\mathrm{base}$: ranger$\to$perform-before, merchant$\to$drink-before, captain$\to$perform-after.
$|\Pi|{=}4$.
\end{sloppypar}

\subsection{Summary Tables}

\begin{table}[h]
\centering
\small
\resizebox{\linewidth}{!}{%
\begin{tabular}{lcccccc}
\toprule
Task & $|V_1|$ & $|V_2|$ & Output space & $|\text{Source}|$ & $|\text{Gen}|$ \\
\midrule
\textit{A-Add} (Additive)      & 3 & 3 & 5 sizes                    & 6 & 3 \\
\textit{A-Comp} (Compositional) & 3 & 3 & $3{\times}3{=}9$ items      & 6 & 3 \\
\textit{A-Cond} (Conditional)   & 4 & 3 & $3{\times}4{=}12$ items     & 8 & 4 \\
\textit{A-Over} (Override)      & 3 & 3 & 4 sizes (3 base + 1 override) & 7 & 2 \\
\bottomrule
\end{tabular}%
}
\caption{Attribute induction instantiation summary.
Output space follows from $f$'s structure: additive gives $|V_1|{+}|V_2|{-}1{=}5$;
compositional gives $|V_1|{\times}|V_2|{=}9$;
conditional gives $|W_\text{color}|{\times}|W_\text{size}|{=}3{\times}4{=}12$ Cartesian combinations (6 achievable under the regime structure);
override gives 3 base sizes plus 1 override.}
\label{tab:prop_tasks}
\end{table}

\begin{table}[h]
\centering
\small
\resizebox{\linewidth}{!}{%
\begin{tabular}{lccccc}
\toprule
Task & $|V_1|$ & $|V_2|$ & $|\Pi|$ & $|\text{Source}|$ & $|\text{Gen}|$ \\
\midrule
Proc1 (Additive)      & 2 & 2 & 3 & 3 & 1 \\
Proc2 (Compositional) & 2 & 2 & 4 & 3 & 1 \\
Proc3 (Conditional)   & 4 & 2 & 3 & 7 & 1 \\
Proc4 (Override)      & 3 & 4 & 4 & 9 & 3 \\
\bottomrule
\end{tabular}%
}
\caption{Procedural induction instantiation summary.
$|\Pi|$ is the number of distinct process templates: additive gives 3 count-variants;
compositional gives $2{\times}|O|{=}2{\times}2{=}4$; conditional gives 3
(two regime-specific $+$ one shared); override gives 3 base $+$ 1 override $=4$.}
\label{tab:proc_tasks}
\end{table}

\section{Experiment Setup Details}
\label{appendix:eval_setup}

\noindent\textbf{Variants.}~For each task type, 20 variants are generated by randomly sampling surface names (entity names, attribute values, item names) from a fixed lexicon and varying the source-gen splits while satisfying the identifiability condition. All metrics are reported as means across these 20 variants.

\noindent\textbf{Prompt structure.}~Each episode prompt contains the world listing, source demonstrations interleaved with $D{=}4$ distractor trajectories, and the gen-split goal.

\noindent\textbf{Distractors.}~Each episode includes $D{=}4$ distractor trajectories interleaved with the source demonstrations. Distractors carry different attributes and arbitrary item assignments with no underlying pattern, requiring the agent to identify which demonstrations are signal-bearing.

\noindent\textbf{Inference settings.}~Table~\ref{tab:inference} lists the decoding hyperparameters used for each evaluated model.

\begin{table}[h]
\centering
\small
\begin{tabular}{lccc}
\toprule
Model & Temp. & Top-$p$ & Max tokens \\
\midrule
Qwen3.5-27B      & 1.0 & 0.95 & 512 \\
Olmo3.1-32B      & 1.0 & 0.95 & 512 \\
GPT-OSS-120B     & 1.0 & 0.95 & 512 \\
Llama-4-Maverick & 1.0 & 0.95 & 512 \\
Gemini-3.1-Flash & 0.1 & —    & 512 \\
GPT-5.4-mini     & —   & —    & 512 \\
\bottomrule
\end{tabular}
\caption{Inference hyperparameters for evaluated models.}
\label{tab:inference}
\end{table}

\noindent\textbf{Computing resources} Models with fewer than 40B parameters were hosted on 2 NVIDIA A40 GPUs. Larger models (e.g., GPT-OSS-120B) were hosted on 1-2 NVIDIA GH200 GPUs. Proprietary models (GPT-5.4-mini, Gemini-3.1-Flash) were accessed via API. Depending on the model, the eight tasks (20 variants each) require approximately 4–7 GPU hours on two NVIDIA A40 GPUs, or 3–5 hours on a GH200.

\noindent\textbf{Episode budget.}~The per-episode action budget is $\text{ref\_length} \times n_\text{tries}$ total actions within one continuous episode, where $\text{ref\_length}$ is the ground-truth solution length and $n_\text{tries}$ is the number of distinct items or processes a brute-force model would need to enumerate to guarantee success. $n_\text{tries}$ is set per task type as follows:

\begin{table}[h]
\centering
\resizebox{\linewidth}{!}{
\begin{tabular}{lcc}
\toprule
Task & $n_\text{tries}$ & Brute-force strategy \\
\midrule
A-Add & $5$ & enumerate item sizes \\
A-Comp & $9$ & enumerate (size, color) pairs \\
A-Cond & $6$ & enumerate per-regime items \\
A-Over & $4$ & enumerate items + exception \\
P-Add & $3$ & enumerate action counts \\
P-Comp & $4$ & enumerate (action, ordering) pairs \\
P-Cond & $3$ & enumerate per-regime processes \\
P-Over & $4$ & enumerate processes + exception \\
\bottomrule
\end{tabular}}
\caption{Per-task $n_\text{tries}$ values and the corresponding brute-force enumeration strategy.}
\end{table}

\noindent\textbf{Budget calculation.}~One \emph{try} is defined as $\text{ref\_length}$: the number of actions in one successful task completion.
For attribute tasks ref\_length$=4$ (the fixed base template); for procedural tasks it varies by entity depending on how many action nodes the hidden rule inserts.
The total episode budget $\text{ref\_length} \times n_\text{tries}$ therefore affords exactly $n_\text{tries}$ complete attempts.
The environment enforces one item acquisition per attempt---the agent cannot buy multiple candidate items in a single step---so a wrong choice consumes a full ref\_length and the attempt counter $t$ in norm\_eff directly counts how many tries were used before success.

\noindent\textit{Example (A-Add, ref\_length$=4$, $n_\text{tries}{=}5$, budget$=20$ actions):}
\begin{quote}\small
Try 1: \texttt{go}(armory) $\to$ \texttt{buy}(size-1 sword) $\to$ \texttt{go}(forest) $\to$ \texttt{defeat}(guardian) \hfill [fail]\\
Try 2: \texttt{go}(armory) $\to$ \texttt{buy}(size-2 sword) $\to$ \texttt{go}(forest) $\to$ \texttt{defeat}(guardian) \hfill [fail]\\
Try 3: \texttt{go}(armory) $\to$ \texttt{buy}(size-3 sword) $\to$ \texttt{go}(forest) $\to$ \texttt{defeat}(guardian) \hfill [success, $t{=}3$]\\
$\Rightarrow$ norm\_eff $= (5/3-1)/(5-1) \approx 0.42$
\end{quote}
A model that induces the rule correctly succeeds on try~1 (norm\_eff$=1.0$); a brute-force enumerator that exhausts all tries achieves norm\_eff$=0$.

\noindent For A-Cond, the full output space contains 12 Cartesian (size, color) combinations, but only 6 are achievable under the conditional regime structure (3 size options in regime-0, 3 color options in regime-1).
Since the gen entity's class is withheld from source, its regime assignment must itself be induced and is unknown to a brute-force agent; such an agent must therefore enumerate all 6 achievable outputs across both regimes, giving $n_\text{tries}{=}6$.

\clearpage

\begin{table*}[b]
  \centering
  \footnotesize
  \setlength{\tabcolsep}{2pt}
  \begin{tabular}{lrrrrrrrrrrrrrrrr}
  \toprule & \multicolumn{2}{c}{A-Add} & \multicolumn{2}{c}{A-Comp} & \multicolumn{2}{c}{A-Cond} & \multicolumn{2}{c}{A-Over} & \multicolumn{2}{c}{P-Add} & \multicolumn{2}{c}{P-Comp} & \multicolumn{2}{c}{P-Cond} & \multicolumn{2}{c}{P-Over} \\
  \cmidrule(lr){2-3} \cmidrule(lr){4-5} \cmidrule(lr){6-7} \cmidrule(lr){8-9} \cmidrule(lr){10-11} \cmidrule(lr){12-13} \cmidrule(lr){14-15} \cmidrule(lr){16-17}
   Model & ECSR & RV & ECSR & RV & ECSR & RV & ECSR & RV & ECSR & RV & ECSR & RV & ECSR & RV & ECSR & RV \\
  \midrule
  Qwen3.5-27B      & 0.22 & 0.12 & 0.42 & 0.30 & 0.29 & 0.03 & 0.33 & 0.07 & 0.41 & 0.00 & 0.01 & 0.00 & 0.26 & \cellcolor{gray!15}0.03 & 0.17 & 0.03 \\
  Gemini-3.1-Flash & 0.49 & \cellcolor{gray!15}0.47 & 0.45 & 0.47 & 0.29 & 0.00 & 0.60 & 0.00 & \cellcolor{gray!15}0.48 & 0.10 & 0.00 & 0.03 & 0.05 & 0.00 & 0.18 & 0.00 \\
  GPT-OSS-120B     & \cellcolor{gray!15}0.85 & \cellcolor{gray!15}0.47 & 0.91 & \cellcolor{gray!15}0.80 & \cellcolor{gray!15}0.38 & \cellcolor{gray!15}0.20 & 0.70 & \cellcolor{gray!15}0.42 & 0.45 & \cellcolor{gray!15}0.45 & 0.03 & \cellcolor{gray!15}0.42 & \cellcolor{gray!15}0.26 & 0.00 & \cellcolor{gray!15}0.47 & \cellcolor{gray!15}0.53 \\
  GPT-5.4-mini     & 0.65 & \cellcolor{gray!15}0.47 & \cellcolor{gray!15}0.93 & 0.60 & 0.35 & 0.00 & \cellcolor{gray!15}0.83 & \cellcolor{gray!15}0.42 & 0.35 & 0.42 & \cellcolor{gray!15}0.12 & 0.07 & 0.16 & 0.00 & 0.21 & 0.30 \\
  Llama-4          & 0.13 & 0.03 & 0.58 & 0.20 & 0.30 & 0.00 & 0.48 & 0.00 & 0.42 & 0.00 & 0.00 & 0.00 & 0.00 & 0.00 & 0.04 & 0.10 \\
  Olmo3.1-32B-Instruct & 0.36 & 0.05 & 0.20 & 0.00 & 0.23 & 0.00 & 0.25 & 0.00 & 0.31 & 0.00 & 0.01 & 0.00 & 0.00 & 0.00 & 0.05 & 0.05 \\
  \rowcolor{yellow!12}
  Humans (avg.)    & 0.63 & 0.71 & 0.75 & 0.79 & 0.57 & 0.64 & 0.78 & 1.00 & 0.61 & 0.71 & 0.57 & 0.57 & 0.62 & 0.57 & 0.58 & 0.71 \\
  \bottomrule
  \end{tabular}
  \caption{Efficiency-calibrated success rate (ECSR) and rule verbalization score (RV, normalised to 0--1) per model and task.}
  \label{tab:main_results}
\end{table*}

\begin{figure*}[t]
  \centering
  \includegraphics[width=\textwidth]{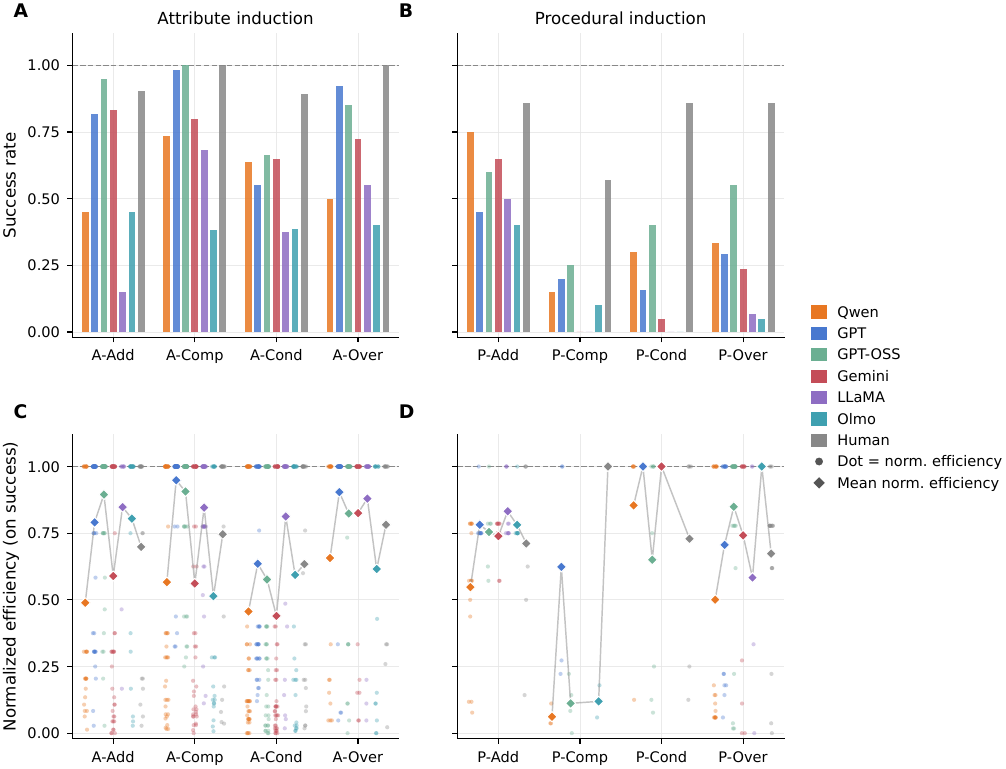}
  \caption{Task success rate (left) and normalized efficiency on successful episodes (right) per model and task, with human baselines shown as grey diamonds.}
  \label{fig_success_efficiency}
\end{figure*}

\section{Full Results}
\label{appx:full_results}

Table~\ref{tab:main_results} reports ECSR and RV for all evaluated models across all eight tasks; Figure~\ref{fig_success_efficiency} further decomposes success rate and normalized efficiency per model. Human performance (bottom row of Table~\ref{tab:main_results}) is shown for comparison.

\section{Human Performance Analysis}
\label{appx:human_analysis}

Human data shows a bimodal pattern: when participants identify the rule (high RV), ECSR is high; when they do not, ECSR drops, reflecting that participants who miss the rule generally cannot complete the task efficiently. Notably, high RV does not always yield optimal ECSR. Decomposing ECSR (Figure~\ref{fig_success_efficiency}) reveals that humans generally succeed but accumulate extra actions on successful episodes---through mis-typing argument names, navigating to the wrong location, or briefly experimenting with the environment. This is not a failure of rule understanding: normalized efficiency exceeds 0.5 on most tasks (i.e., succeeding within roughly the second attempt), far from brute-force enumeration.

\section{Experiment Analysis}
\label{appx:additional}

\begin{figure*}[t]
  \centering
  \includegraphics[width=\textwidth]{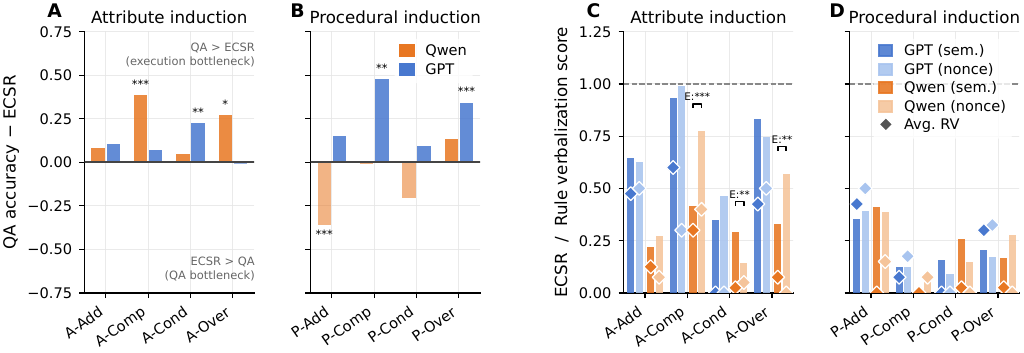}
  \caption{
  (A--B) Format gap (QA accuracy $-$ ECSR) per model and task, for attribute (A) and procedural (B) tasks.
  Bars above zero indicate an \emph{execution bottleneck}; bars below zero a \emph{QA bottleneck}.
  Stars mark gaps significantly non-zero (Bonferroni-corrected by task family;
  $^{*}p<0.05$, $^{**}p<0.01$, $^{***}p<0.001$).
  (C--D) ECSR and RV under semantic vs.\ nonce conditions for attribute (C) and procedural (D) tasks.
  Brackets mark significant pairwise differences (Bonferroni-corrected by task family}
  \label{fig:gap_nonce_combined}
\end{figure*}

\subsection{Does Surface Semantics Help?}
\label{appx:nonce}

\begin{table}[htb]
\centering
\small
\caption{Effect of lexical condition ($\Delta\mu = \text{semantic} - \text{nonce}$) on ECSR and RV, aggregated by task family.
\colorbox{blue!15}{Blue}: significant advantage.}
\label{tab:nonce}
\begin{tabular*}{\columnwidth}{@{\extracolsep{\fill}} l r r r r}
\toprule
& \multicolumn{2}{c}{\textbf{ECSR}} & \multicolumn{2}{c}{\textbf{RV}} \\
\cmidrule(lr){2-3}\cmidrule(lr){4-5}
\textbf{Models} & \textbf{Attr.} & \textbf{Proc.} & \textbf{Attr.} & \textbf{Proc.} \\
\midrule
GPT-5.4-mini        & $-0.015$\hphantom{\textsuperscript{**}} & $\phantom{-}0.017$ & $\phantom{-}0.050$ & $-0.050$ \\
Qwen3.5-27B  & \cellcolor{blue!15}$-0.126$\textsuperscript{**} & $-0.008$ & $\phantom{-}0.000$ & $-0.044$ \\
\bottomrule
\end{tabular*}
\end{table}

Surface semantics could in principle affect performance if models leverage
pretraining knowledge to constrain the rule space, for instance by
associating entity names or attribute values with plausible item choices (e.g., a weapon is needed to defeat a guardian).
We test this by comparing ECSR and RV under two lexical conditions:
\emph{semantic}, where all names are drawn from a natural-language lexicon,
and \emph{nonce}, where they are replaced by meaningless syllables
(e.g., \texttt{vrel\_A}, \texttt{korv\_Z}), eliminating any surface cue
that could be exploited via world knowledge.

\noindent\textbf{Results.}~Table~\ref{tab:nonce} shows ECSR and RV aggregated by task family under both conditions
(sign-permutation test, $5{,}000$ permutations, two-sided; per-task breakdown in
Figure~\ref{fig:gap_nonce_combined} above).
GPT-5.4-mini is virtually insensitive to lexical condition across both task families:
ECSR $\Delta\mu = 0.001$ ($p = 0.97$) and RV $\Delta\mu = 0.000$ ($p = 1.0$).
Qwen3.5-27B shows no significant difference on procedural tasks ($p = 0.83$),
but a small ECSR advantage for nonce over semantic on attribute tasks
($\Delta\mu = {-}0.126$, $p = 0.02$, Bonferroni-corrected by task family). This is not surprising given the benchmark design: rules are deliberately constructed without real-world associations (e.g., a large entity may require a small weapon), so semantic names provide no reliable advantage and may actively mislead.

\begin{figure}[htb]
  \centering
  \includegraphics[scale=1]{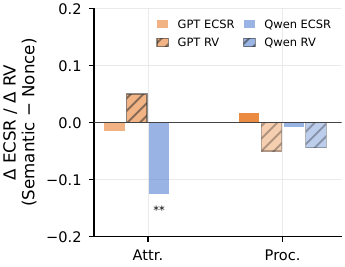}
  \caption{ECSR and RV for GPT-5.4-mini and Qwen3.5-27B under semantic vs.\ nonce
  lexical conditions, for attribute (left) and procedural (right) tasks.
  Brackets mark significant pairwise differences (Bonferroni-corrected by
  task family; $p<0.05$); unmarked pairs are non-significant.}
  \label{fig:nonce}
\end{figure}

\subsection{Do Existing Methods Improve Task Performance?}
\label{appx:methods}
\label{appx:steering-method-performance}

A growing body of work has proposed methods to improve LLMs' inductive reasoning \citep{qiu-etal-2024-phenomenal, he2024idea, lee-etal-2025-moc, chen2025survey}. We evaluate a representative subset on \AS: two induction-specific methods and two general reasoning strategies, applied to both GPT-5.4-mini and Qwen3.5-27B.
The two general methods are \textit{ReAct} \citep{yao2023react}, which interleaves chain-of-thought with action steps, and \textit{ACE} \citep{zhang2026agentic}, which selects and orders demonstrations to maximally inform rule induction. The two induction-specific methods are \textit{Hypothesis Refinement} (HR; \citealt{qiu-etal-2024-phenomenal}), which iteratively refines a candidate rule against counterexamples, and \textit{IDEA} \citep{he2024idea}, which cycles through induction, deduction, and abduction steps. Prompts are in Appx.~\ref{appx:prompts}.

\begin{figure}[htb]
  \centering
  \includegraphics[scale=1]{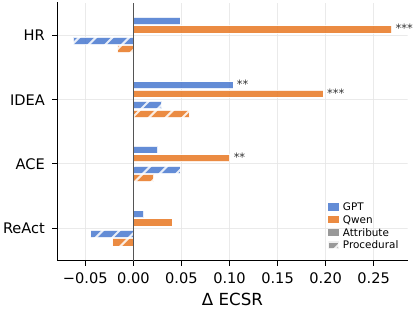}
  \caption{ECSR change relative to the GPT-5.4-mini and Qwen3.5-27B baselines ($\Delta$ECSR)
  under different methods (HR, IDEA, ACE, ReAct), per task.
  Stars mark significant $\Delta$ECSR ($p$-values; $^{*}p<0.05$, $^{**}p<0.01$,
  $^{***}p<0.001$; Bonferroni-corrected by task family).
  Full numerical results in Table~\ref{tab:phase2_results}.}
  \label{fig:improvement}
\end{figure}

\noindent\textbf{Results.}~Aggregating across all tasks $\times$ models (Figure~\ref{fig:improvement}), IDEA yields
the largest overall gain ($\Delta\mu = 0.097$, 95\% CI $[0.062,\,0.131]$,
$p < 0.001$; Bonferroni-corrected), followed by HR ($\Delta\mu = 0.059$, $p = 0.004$) and ACE
($\Delta\mu = 0.049$, $p = 0.010$); ReAct provides no reliable benefit
($\Delta\mu = {-}0.004$, $p = 0.82$).
Significant gains are concentrated on attribute tasks (Figure~\ref{fig:improvement}); no method yields reliable improvement on procedural tasks, though IDEA shows a consistent positive trend there.
HR and IDEA are more reliable and broadly effective than ACE and ReAct, with ReAct the only method showing significant degradation on some tasks, suggesting induction-specific methods outperform general reasoning strategies.

\begin{table*}[htb]
  \centering
  \footnotesize
  \setlength{\tabcolsep}{4pt}
  \begin{tabular*}{\textwidth}{@{\extracolsep{\fill}} l rrrr rrrr}
  \toprule
  & \multicolumn{4}{c}{\textbf{Attribute}} & \multicolumn{4}{c}{\textbf{Procedural}} \\
  \cmidrule(lr){2-5}\cmidrule(lr){6-9}
  Method & \textit{A-Add} & \textit{A-Comp} & \textit{A-Cond} & \textit{A-Over} & \textit{P-Add} & \textit{P-Comp} & \textit{P-Cond} & \textit{P-Over} \\
  \midrule
  GPT-5.4-mini                              & 0.65 & 0.93 & 0.35 & 0.83 & 0.35 & 0.12 & \cellcolor{gray!15}0.16 & 0.21 \\
  \quad + HR \citep{qiu-etal-2024-phenomenal}   & 0.75 & 0.97 & 0.44 & 0.79 & 0.24 & 0.09 & 0.06 & 0.20 \\
  \quad + IDEA \citep{he2024idea}               & \cellcolor{gray!15}$0.82^{**}$ & \cellcolor{gray!15}0.99 & \cellcolor{gray!15}$0.50^{**}$ & \cellcolor{gray!15}0.86 & 0.30 & \cellcolor{gray!15}0.20 & 0.11 & \cellcolor{gray!15}0.34 \\
  \quad + ACE \citep{zhang2026agentic}          & 0.68 & 0.89 & $0.49^{*}$ & 0.79 & \cellcolor{gray!15}0.44 & 0.19 & 0.13 & 0.27 \\
  \quad + ReAct \citep{yao2023react}            & 0.67 & 0.99 & 0.37 & 0.77 & 0.27 & 0.12 & 0.10 & 0.17 \\
  \midrule
  Qwen3.5-27B                               & 0.22 & 0.42 & 0.29 & 0.33 & 0.41 & 0.01 & 0.26 & 0.17 \\
  \quad + HR \citep{qiu-etal-2024-phenomenal}   & \cellcolor{gray!15}$0.41^{**}$ & \cellcolor{gray!15}$0.87^{***}$ & \cellcolor{gray!15}$0.40^{*}$ & \cellcolor{gray!15}$0.65^{***}$ & 0.30 & \cellcolor{gray!15}0.14 & 0.14 & 0.21 \\
  \quad + IDEA \citep{he2024idea}               & 0.35 & $0.80^{***}$ & 0.34 & $0.56^{**}$ & \cellcolor{gray!15}0.43 & $0.06^{*}$ & \cellcolor{gray!15}0.34 & 0.24 \\
  \quad + ACE \citep{zhang2026agentic}          & 0.27 & 0.50 & 0.33 & $0.55^{**}$ & 0.40 & 0.04 & 0.24 & 0.24 \\
  \quad + ReAct \citep{yao2023react}            & $0.12^{*}$ & 0.55 & 0.27 & 0.47 & $0.28^{*}$ & 0.02 & 0.20 & \cellcolor{gray!15}0.26 \\
  \bottomrule
  \end{tabular*}
  \caption{ECSR under different induction methods applied to GPT-5.4-mini (top) and Qwen3.5-27B (bottom).
  Stars denote significant $\Delta$ECSR vs.\ baseline ($^{*}p<0.05$, $^{**}p<0.01$, $^{***}p<0.001$; raw $p$-values).
  Shaded cells indicate the best value per column across both model groups.
  HR = Hypothesis Refinement; IDEA = Induction--Deduction--Abduction.}
  \label{tab:phase2_results}
\end{table*}

\clearpage

\section{LLM-Judge Ground Truth}
\label{appx:llm-judge-ground-truth}
\codeboxinput[label={lst:gt}]{LLM-judge ground truth}{prompts/ground_truth.txt}
\clearpage
\section{Prompts}
\label{appx:prompts}

\subsection{Main Task Prompt}
\label{appx:main-task-prompt}
\codeboxinput[label={lst:base-prompt}]{Main Task Prompt}{prompts/main_task_prompt.txt}

\subsection{Rule Verbalization}
\label{appx:rule-verbalization-prompts}
\codeboxinput[label={lst:rv-prompt}]{RV Prompt}{prompts/rule_verbalization.txt}
\codeboxinput[label={lst:rv-judge}]{RV Judge Prompt}{prompts/rule_score_llm_judge.txt}

\subsection{QA Prompt}
\label{appx:qa-prompts}
\codeboxinput[label={lst:qa-judge}]{Attribute Induction QA Prompt}{prompts/QA_attr.txt}
\codeboxinput[label={lst:qa-proc-judge}]{Procedural Induction QA Prompt}{prompts/QA_proc.txt}

\subsection{Steering Methods Prompts}
\label{appx:steering-methods-prompts}
\codeboxinput[label={lst:steering-methods-prompts}]{ReAct}{prompts/react.txt}
\codeboxinput[label={lst:steering-methods-prompts}]{Hypothesis Refinement}{prompts/hr_hypothesis.txt}
\codeboxinput[label={lst:steering-methods-prompts}]{IDEA Abduction}{prompts/idea_abduction.txt}
\codeboxinput[label={lst:steering-methods-prompts}]{IDEA Induction}{prompts/idea_induction.txt}
\codeboxinput[label={lst:steering-methods-prompts}]{ACE Reflector}{prompts/ace_reflector.txt}
\codeboxinput[label={lst:steering-methods-prompts}]{ACE Deduplication}{prompts/ace_deduplication.txt}

\section{Human annotation interface}
\label{appx:interface}
\begin{figure*}
    \centering
    \includegraphics[width=\textwidth]{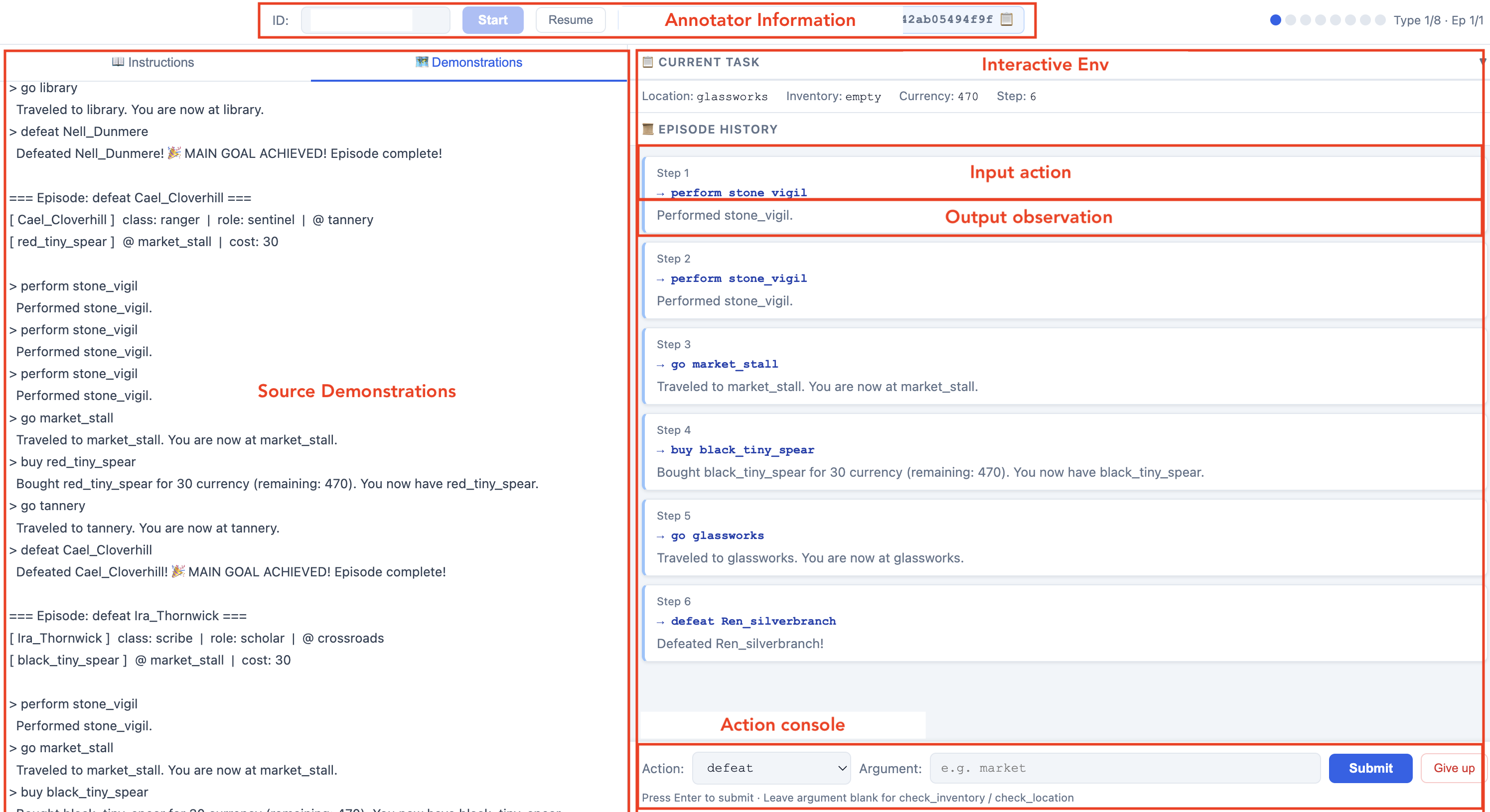}
    \caption{Episodic environment interface for humans to play the games; We use the same instructions as the one we feed to models see Appx. \ref{appx:main-task-prompt}.}
    \label{fig:interface}
\end{figure*}

\begin{figure*}
    \centering
    \includegraphics[width=\textwidth]{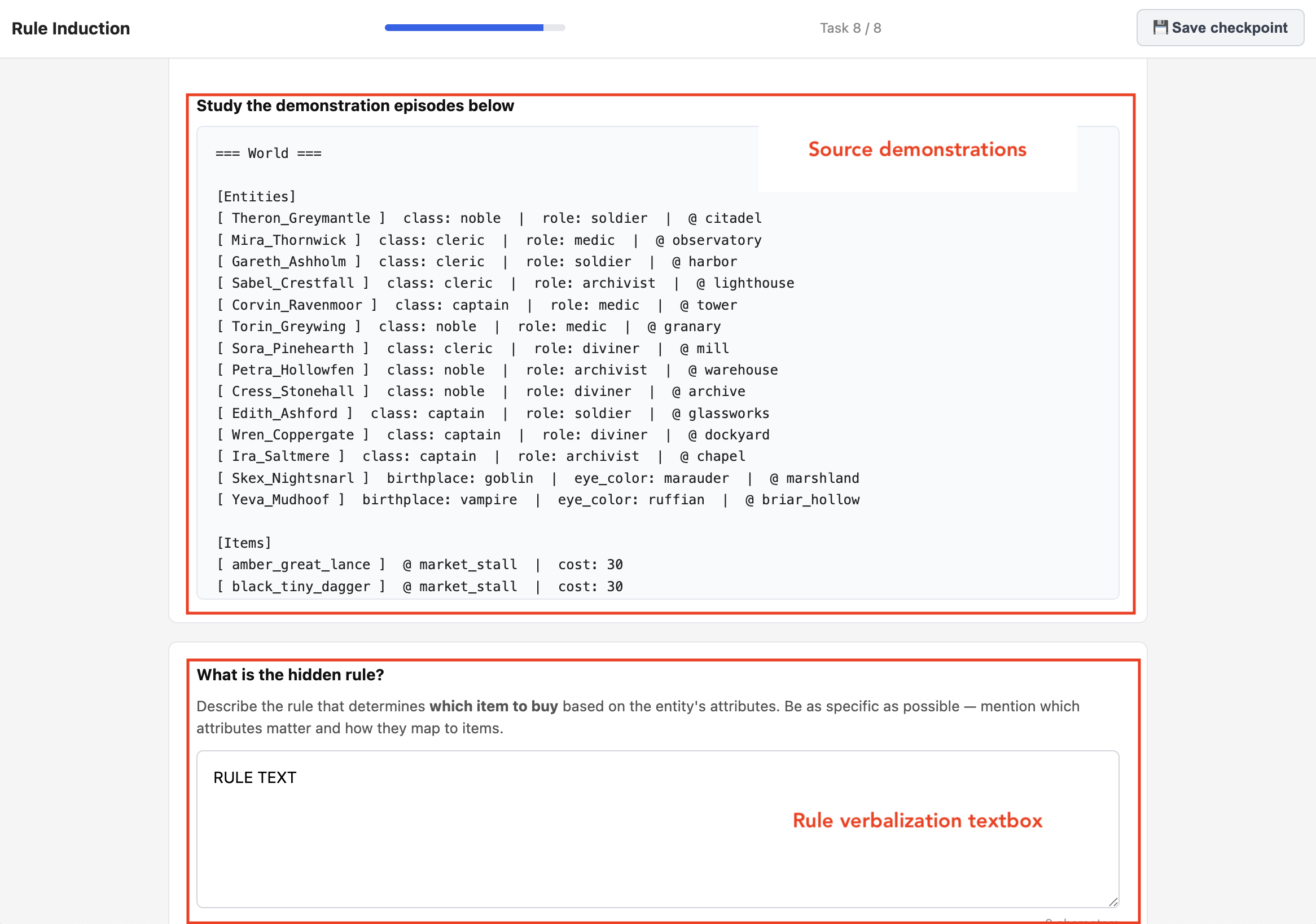}
    \caption{Interface to annotate the rule underlying the demonstrations after each episode}
    \label{fig:interface2}
\end{figure*}
\end{document}